\documentclass{article}

\usepackage[utf8]{inputenc}
\usepackage[T1]{fontenc}
\usepackage{lmodern}
\usepackage[figurename=Figure,labelfont=bf,labelsep=period]{caption}
\usepackage{subcaption}
\usepackage{amsmath}

\usepackage{newtxtext,newtxmath}
\usepackage[colorlinks=true,citecolor=red,linkcolor=black]{hyperref}

\usepackage{soul}
\usepackage{array}
\usepackage{balance}
\usepackage{longtable}
\usepackage[T1]{fontenc}
\usepackage[flushleft]{threeparttablex}
\usepackage[symbol*]{footmisc}
\usepackage{booktabs}
\usepackage{pifont}
\usepackage{soul}
\usepackage{algorithm}
\usepackage{algpseudocode}
\usepackage{svg}
\usepackage{graphicx}
\usepackage{multirow}
\usepackage{booktabs}

\newsavebox\mybox

\usepackage{etoolbox}
\usepackage{comment}
\usepackage{booktabs}
\usepackage{makecell}
\usepackage{threeparttable}
\usepackage{subcaption}

\let\svthefootnote\thefootnote
\newcommand\freefootnote[1]{%
  \let\thefootnote\relax%
  \footnotetext{#1}%
  \let\thefootnote\svthefootnote%
}

\usepackage{arxiv}
\usepackage[utf8]{inputenc} 
\usepackage[T1]{fontenc}    
\usepackage{hyperref}       
\usepackage{url}            
\usepackage{amsfonts}       
\usepackage{nicefrac}       
\usepackage{microtype}      
\usepackage{lipsum}
\usepackage{graphicx}
\graphicspath{ {./images/} }

\title{Decoding Psychological States Through Movement: Inferring Human Kinesic Functions with Application to Built Environments}

\author{
 Cheyu Lin \\
  Dept. of Civil \& Environmental Engineering\\
  Carnegie Mellon University\\
  Pittsburgh, PA 15213 \\
  \texttt{cheyul@andrew.cmu.edu} \\
   \And
 Katherine A. Flanigan$^*$ \\
  Dept. of Civil \& Environmental Engineering\\
  Carnegie Mellon University\\
  Pittsburgh, PA 15213 \\
  \texttt{kflaniga@andrew.cmu.edu} \\
  $^*$Corresponding Author
  \And
 Sirajum Munir \\
  Bosch Research and Technology Center\\
  Pittsburgh, PA 15222 \\
  \texttt{sirajum.munir@us.bosch.com}
}
\date{}
\begin{document}
\maketitle

\widowpenalty=0
\clubpenalty=0
\flushbottom

\begin{abstract}
Social infrastructure and other built environments are increasingly expected to support belonging, well-being, and community resilience by enabling social interaction. Yet in civil and built-environment research, there is no consistent, scalable, and privacy-preserving way to represent and measure socially meaningful interaction in these spaces, leaving studies to operationalize “interaction” differently across contexts and limiting practitioners’ ability to evaluate whether design interventions are changing the forms of interaction that social capital theory predicts should matter. To address this coupled field-level and methodological gap, we introduce the \mbox{\textbf{D}}yadic \mbox{\textbf{U}}ser \mbox{\textbf{E}}ngagemen\mbox{\textbf{T}} (DUET) dataset and an embedded kinesics recognition framework that operationalize Ekman and Friesen’s kinesics taxonomy as a function-level interaction vocabulary aligned with social capital-relevant behaviors (e.g., reciprocity, attention coordination, and turn-taking). DUET captures 12 day-to-day dyadic interactions spanning all five kinesic functions—emblems, illustrators, affect displays, adaptors, and regulators—across four sensing modalities (RGB, infrared, depth, and 3D skeletal keypoints) and three built-environment contexts, enabling privacy-preserving analysis of communicative intent through movement. Benchmarking six open-source, state-of-the-art human activity recognition models quantifies the difficulty of communicative-function recognition on DUET and highlights the limitations of ubiquitous monadic, action-level recognition when extended to dyadic, socially grounded interaction measurement. Building on DUET, our recognition framework infers communicative function directly from privacy-preserving skeletal motion without handcrafted action-to-function dictionaries; using a transfer-learning architecture, it reveals structured clustering of kinesic functions and a strong association between representation quality and classification performance ($\rho=0.91$, 95\% CI [0.82, 0.96]) while generalizing across subjects and contexts. Together, DUET and the proposed framework provide measurement and inference infrastructure that operationalizes an interpretable, computationally measurable “interaction layer” for evaluating social infrastructure interventions and enabling future closed-loop, evidence-based design of built environments that promote social connection and social capital.
\end{abstract}

\section{Introduction} \label{sec:Introduction} \vspace{-0.2cm} 
\sloppy

\subsection{Social Interaction as an Emergent Property of the Built Environment}

The built environment plays a critical role in shaping human psychology, cognition, and social behavior. This insight is far from new: since the 1970s, space syntax theorists have demonstrated that the spatial configuration of the built environment systematically shapes patterns of movement, encounter, and co-presence \mbox{\cite{hillier_space_1976,hillier_social_1989,hillier2007space}}. Through measures such as integration and visual connectivity, this body of work established the principle that spatial form exerts a generative influence on human behavior. More recently, disciplines such as psychology, sociology, architecture, and computational behavior modeling (e.g., reinforcement learning) have expanded this viewpoint by examining how spatial and environmental features shape human activity \mbox{\cite{schneider2002school,hillier2007space,brown_architecture_2014,koutsolampros_dissecting_2019,mccunn2021lighting,sander2025natural}}. In urban public spaces, for example, seating configurations, street-level visibility, and spatial openness have been shown to influence where people congregate, how long they linger, and how frequently spontaneous social interactions occur \mbox{\cite{whyte_social_1980,gehl_life_1987}}. Similarly, studies in educational environments demonstrate that school layout and design meaningfully influence student behavior, engagement, and peer interaction patterns, with features such as openness, circulation structure, and spatial accessibility shaping how and where students connect \mbox{\cite{schneider2002school}}. Taken together, these findings illustrate that design choices in the built environment can increase or suppress opportunities for co-presence and interaction—an essential precursor to human-centered outcomes.

Understanding such built environment-behavior relationships has taken on new urgency as mounting evidence demonstrates the societal importance of \textit{social infrastructure} \mbox{\cite{aldrich_building_2012,klinenberg_palaces_2018,murthy_our_2023,fraser_tale_2024}}---defined across civil engineering as the built environment spaces and environmental affordances that enable, mediate, and sustain social interactions among community members \mbox{\cite{elrod_2025_2}}. Social infrastructure such as libraries, parks, community centers, and workplace environments have been linked to critical societal outcomes including disaster mortality reduction \mbox{\cite{aldrich_building_2012}}, sense of community \mbox{\cite{francis_2012}}, and public health across multiple dimensions \mbox{\cite{fraser_better_2022,fraser_great_2024}}. These benefits are widely attributed to the accumulation of \textit{social capital}---the capacity of human networks to mobilize information, resources, and support through patterns of interaction grounded in trust, reciprocity, and shared norms \mbox{\cite{putnam_bowling_2000,aldrich_building_2012,klinenberg_palaces_2018,murthy_our_2023,fraser_tale_2024}}. Communities with higher levels of social capital experience lower mortality and improved health outcomes, recover more quickly from disasters, and demonstrate greater civic participation, to name a few \mbox{\cite{aldrich_building_2012,aldrich_social_2015,murthy_our_2023,aldrich_how_2023,fang_brewing_2024,damore_community_2025}}.

\begin{figure*}[!t]
  \centering
  \begin{subfigure}[b]{1\linewidth}
  \centering
    \includegraphics[width=1\linewidth]{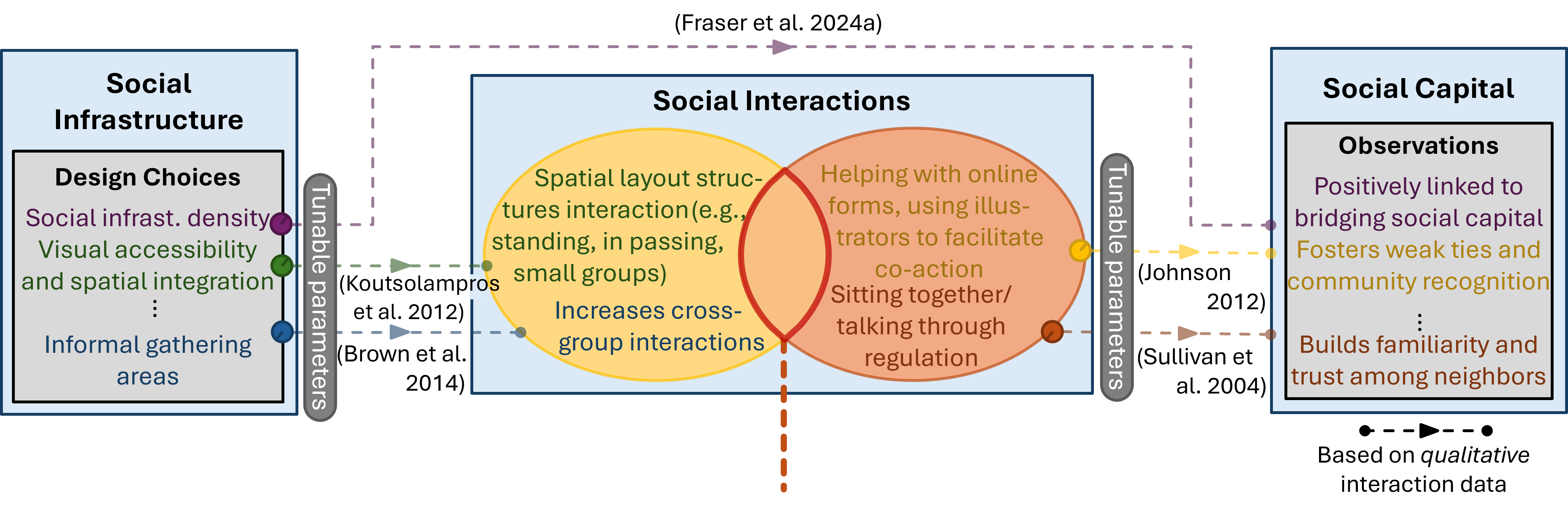}
    \vspace{-0.8cm} \caption{}
    \label{fig:Figure0a}
  \end{subfigure}\\
  \begin{subfigure}[b]{1\linewidth}
  \centering
    \includegraphics[width=1\linewidth]{Images/Figure1b.png}
    \caption{}
    \label{fig:Figure0b}
  \end{subfigure}%
  \caption{(a) Existing literature provides three separate sets of evidence: (i) studies showing that social infrastructure shapes social interactions (Social Infrastructure $\rightarrow$ yellow oval), (ii) studies showing that social interactions contribute to social capital (orange oval $\rightarrow$ Social Capital), and (iii) studies correlating social infrastructure directly with social capital (Social Infrastructure $\rightarrow$ Social Capital). However, no existing work connects these components through a shared, measurable representation of the interaction layer, outlined in red.
(b) This paper addresses that gap by introducing a quantitative, kinesics-based approach for observing the interaction mechanisms that mediate the effects of spatial design on social capital, enabling a systematic pathway from social infrastructure to interactional function to social capital.}
  \label{fig:Figure0} \vspace{-0.4cm}
\end{figure*}

However, \mbox{\textit{how}} social infrastructure gives rise to social capital remains insufficiently understood. Recent literature increasingly indicates that the pathway operates through the \mbox{\textit{social interactions}} that physical spaces facilitate \mbox{\cite{johnson_how_2012}}. The physical environment provides affordances---opportunities for encounters, visibility between potential interaction partners, spaces conducive to lingering---that increase the likelihood, frequency, and quality of social interactions \mbox{\cite{leyden_social_2003,williams_how_2019,kuzuoglu_how_2024,morales-flores_understanding_2025}}. These interactions, accumulated over time, build the trust and reciprocity that constitute social capital. Studies across social infrastructure types consistently demonstrate this pathway  \mbox{\cite{johnson_how_2012,small_someone_2017}}. Figure \mbox{\ref{fig:Figure0a}} illustrates this pathway and provides representative examples linking (a) social infrastructure to social interactions and (b) social interactions to social capital.

Despite evidence that social infrastructure shapes the interactions people have, and that social interactions in turn contribute to social capital, this literature remains conceptually disconnected. Both demonstrate that interactions matter, but they treat those interactions as context-specific phenomena---defined, observed, and analyzed differently across studies (Figure \mbox{\ref{fig:Figure0a}}). Accordingly, a concrete domain-driven problem emerges: civil engineering practitioners can alter spatial design and add or retrofit social infrastructure, but they have limited ability to measure—systematically and at scale—whether these interventions are changing the interaction mechanisms that social capital theory predicts should matter, or to use this data to support closed-loop “what-if” scenario modeling. Without a shared representation of what an interaction is at a mechanistic level, the field lacks a way to link built environment design to social outcomes in order to model or design spaces with confidence that specific features will produce intended social outcomes.

To move beyond correlation-driven understanding of social infrastructure, we must be able to observe the interaction layer that links spatial design to social outcomes (Figure \mbox{\ref{fig:Figure0b}}). At present, built environment interventions (e.g., adding parks, modifying layouts) are evaluated primarily through long-term social capital indicators, providing no feedback about the interaction processes through which they may succeed or fail. Without the ability to sense and quantify social interactions using a taxonomy grounded in social capital, the system remains open-loop: as seen in Figure \mbox{\ref{fig:Figure0a}}, we can adjust design inputs in a social infrastructure setting but cannot observe the internal state of interaction dynamics that translates those inputs into outcomes, limiting our capacity to attribute outcomes to specific features or iteratively improve them, as shown in Figure \mbox{\ref{fig:Figure0b}}. Quantitatively measuring interactions provides the state information needed for feedback, enabling assessment of whether a space is fostering the types of encounters that theory suggests matter, supporting evidence-based comparison of design alternatives (``what-if'' evaluation), adapt design decisions accordingly, and ultimately move toward a closed-loop, evidence-based approach to creating environments that support social capital formation. In this work, we argue that the ``interaction layer'' must be operationalized in a way that is both interpretable—so it aligns with social theory—and computationally measurable—so it can scale across built environment contexts.

\vspace{-0.15cm} \subsection{Kinesics as a Functional Lens for Social Interaction} \vspace{-0.1cm}

Existing methods for observing and understanding these dyadic (or two-person) interactions remain limited. Traditional approaches---including theory-driven behavior models (e.g., of planned behavior; \mbox{\cite{ajzen1991theory}}), questionnaire-based methods, and manual observation protocols---either oversimplify the heterogeneity of human experience or offer only static, self-reported snapshots that lack continuity and scalability \mbox{\cite{doctorarastoo2023modeling}}. Participants have been shown to adapt their behavior when they know they are being observed, and self-reports can reflect social desirability bias rather than actual behavior. While these methods have validated correlations between spatial features and behavior in post-occupancy studies, they are labor-intensive, episodic, and unsuitable for systematic, continuous monitoring or longitudinal analysis \mbox{\cite{lin2024your}}.

Overcoming several of these limitations, facial expression analysis has been extensively studied for emotion recognition, but it is limited in its ability to capture the full range of communicative functions humans express through movement \mbox{\cite{finn2024influence,sjoraida2024nonverbal}}. Facial cues often encode affective states---such as happiness, anger, or surprise---but provide limited insight into interactional intent and the dynamics of social coordination needed to capture key social capital indicators \mbox{\cite{adyapady2023}}. Additionally, it is privacy invasive, which is not suitable for monitoring humans in public built environment spaces. A compelling alternative lies in the body. Human experience is not only described---it is enacted. Through posture shifts, gestures, gaze, turn taking, and proximity, people continuously signal their intentions, emotions, and social orientations. These expressive movements, studied under the umbrella of \mbox{\textit{kinesics}}, manifest internal states externally and are foundational to social interaction and psychological regulation \mbox{\cite{ekman1969repertoire}}. Kinesic functions should be understood as correlational rather than deterministic: bodily cues reflect the communicative and regulatory functions through which psychological states are expressed, not the states themselves. They are observable indicators that enable consistent, theory-grounded labeling without overclaiming access to subjective experience. Within a built environment context, this offers a path to define interaction in terms that are both interpretable to social theory and measurable in situ.

To operationalize this insight for computational measurement, we return to a foundational psychological framework: the kinesic taxonomy of Ekman and Friesen \mbox{\cite{ekman1969repertoire}}. The taxonomy classifies bodily movements into five categories based on communicative functions---emblems (intentional symbolic gestures), illustrators (gestures accompanying speech), affect displays (emotional expressions), adaptors (self-regulatory behaviors), and regulators (turn-taking and conversation management)---each reflecting distinct psychological or interpersonal intent. Importantly, these functional distinctions align with social capital theory, which emphasizes that different forms of social connection emerge through different patterns of interaction: for example, brief acknowledgment behaviors support weak-tie formation, while sustained, reciprocal exchanges underpin strong-tie development \mbox{\cite{roberts2015managing,urena2020estimating}}. In this context---described in full in Section \mbox{\ref{sec:Taxonomy}}---kinesic categories serve as interpretable proxies for the behavioral mechanisms through which social capital develops. Specifically, the taxonomy provides an interpretable “interaction vocabulary” that can be used to quantify whether a space is supporting the kinds of reciprocity-, attention-, and coordination-related behaviors, for example, that precede tie formation, trust, and the accumulation of social capital, providing a consistent basis for evaluating whether social infrastructure designs and interventions are fostering interaction patterns that theory predicts should matter. 

\vspace{-0.25cm} \subsection{Proposed Dataset and Kinesics Recognition Framework}

This paper operationalizes this taxonomy in a machine learning context by introducing a dyadic, multi-modal dataset and a privacy-preserving recognition framework that enable the direct measurement of these communicative functions at scale. 
Computational systems today remain largely blind to the psychological meanings of motion, particularly in contexts with more than one human. In human activity recognition (HAR), prevailing models excel at identifying physical actions---walking, waving, sitting---but fail to capture their underlying functions: Why is a person gesturing? What is the intent behind a posture? How is it shaped by the presence of others? Existing models are typically trained on monadic (i.e., single-person) datasets that isolate individuals from social context and lack functional annotations grounded in psychological theory, preventing them from distinguishing between interaction types that social capital theory identifies as consequential. A thorough literature review is provided in Section \mbox{\ref{sec:HAR}}. By grounding HAR in Ekman and Friesen's taxonomy, this work creates a bridge from fine-grained movement capture to the higher-level goal of social interaction intent, which contributes to social capital formation. The developed \mbox{\textbf{D}}yadic \mbox{\textbf{U}}ser \mbox{\textbf{E}}ngagemen\mbox{\textbf{T}} (DUET) dataset and accompanying recognition framework provide the measurement infrastructure required to pursue this goal, enabling automated, scalable, and privacy-preserving inference of communicative function in dyadic interactions. 

Towards this end, this paper overcomes a major technical challenge: the immense variability in human activities makes it infeasible to rely on dictionary-based mappings that exhaustively link actions to corresponding communicative functions. As with questionnaire-based approaches, creating comprehensive mappings would require considerable time and resources, and the result would still be limited in generalizability and adaptability. To enable scalable interpretation of bodily movement, we propose an approach that infers communicative function without handcrafted mappings.

This challenge demands two critical components: (1) a learning framework capable of bypassing the dictionary-based translation process, and (2) a dataset explicitly designed to train and validate such a framework. We position the contributions of this work---introducing the DUET dataset and an embedded kinesics recognition framework---as delivering precisely this foundation. DUET contains 12 dyadic activities spanning all five kinesic communication functions. These activities are intentionally designed to occur between two individuals to capture the reciprocal and context-sensitive nature of nonverbal communication. For example, the gestures waving in and holding up a palm to pause conversation may appear similar when viewed from the initiator alone; however, the responder's behavior---whether they approach or remain distant---provides crucial contextual information to distinguish their meaning. By incorporating both actors in each interaction, DUET enables modeling of communicative function as an emergent property of the dyadic exchange.

DUET offers four sensing modalities---RGB video, depth, infrared (IR), and 3D skeletal keypoints---and is collected across three locations on the Carnegie Mellon University campus (Pittsburgh, PA). These collection sites introduce background variation that helps prevent overfitting to specific visual features, improving model robustness. To ensure spatial consistency across environments, we developed a new data collection protocol that uses a single fixed sensor while systematically varying viewpoints. The resulting dataset comprises 14,400 samples across 12 activities, yielding the highest sample-to-class ratio among publicly available dyadic datasets.

To complement and demonstrate the effectiveness of the dataset, we introduce an embedded kinesics recognition framework that enables function-level inference from bodily movement data without requiring a pre-defined action-to-function dictionary. Our framework leverages a transfer learning architecture that combines a pre-trained spatial-temporal graph convolutional network (ST-GCN) \mbox{\cite{yan2018spatial}}---frozen to preserve low-level motion representations---with a trainable convolutional neural network (CNN) head that learns to cluster activities based on their shared communicative function. Crucially, the framework operates on skeletal keypoint data, which contains no identifying visual features, thereby supporting privacy-preserving inference suitable for human-centered social-infrastructure applications \cite{10.1115/1.4067744}. We implement and evaluate this framework on DUET, demonstrating that kinesic functions form structured, learnable clusters in latent space and that improvements in ST-GCN representation quality yield proportional gains in CNN classification accuracy ($\rho=0.91$, $p = 0.000002$), providing a proof-of-concept for scalable, generalizable, and psychologically informed HAR.

\vspace{-0.2cm} \subsection{Summary of Contributions and Paper Outline}

To summarize the main contributions, this paper targets two coupled gaps that currently limit progress in built environment research on social infrastructure. At the domain level, social infrastructure and social capital literature both emphasize the importance of interaction, yet “interaction” is defined and measured inconsistently across studies, leaving practitioners without a systematic, scalable way to evaluate whether design interventions are changing the forms of interaction that theory predicts should matter. At the methodological level, prevailing human activity recognition systems needed for interaction inference are largely optimized for monadic, action-level classification and therefore struggle to infer the functional meaning of movement—especially in dyadic settings where intent depends on reciprocal context—making them a poor fit for operationalizing an interpretable, privacy-preserving interaction layer in public built environment settings. To address these gaps, we contribute:

\vspace{-0.15cm} \begin{enumerate}
    \item DUET, a dyadic, privacy preserving multimodal dataset grounded in Ekman and Friesen’s kinesics taxonomy—a function-level interaction vocabulary that aligns with social capital-relevant behaviors occurring in built environments such as reciprocity, attention coordination, and turn-taking—to support consistent labeling of everyday interactions across varied built environment contexts; 
    \item Benchmarking of six open-source, state-of-the-art HAR models to quantify the difficulty of recognizing communicative functions on DUET and reveal limitations of widely used monadic, action-level activity recognition paradigms, which do not readily extend to dyadic, socially grounded interaction measurement in social infrastructure settings;
    \item An embedded kinesics recognition framework that infers communicative categories directly from privacy-preserving skeletal motion without static action-to-function dictionaries.
\end{enumerate}

\vspace{-0.15cm} The remainder of this paper is organized as follows. Section \mbox{\ref{sec:Taxonomy}} outlines the kinesics taxonomy of Ekman and Friesen \mbox{\cite{ekman1969repertoire}}, describing the HAR-driven information inferred from each category, linking the taxonomy to social capital, and contextualizing the taxonomy within built environments \cite{lin2025actionskinesicsextractinghuman}. Section \mbox{\ref{sec:HAR}} reviews recent HAR developments and their limitations in modeling communicative function. Section \mbox{\ref{sec:DUET}} introduces the DUET dataset—covering activity types, sensing modalities, acquisition protocols, testbed configurations, data management, subject details, and data partitioning—and notes its availability on Hugging Face \mbox{\cite{huggingfaceAnonymousUploader1DUETDatasets2}}. Section \mbox{\ref{sec:benchmarking}} benchmarks six open-source, state-of-the-art HAR algorithms, demonstrating DUET’s complexity and the limits of current function-level modeling. Section \mbox{\ref{sec:krf}} presents our kinesics recognition framework, which infers communicative categories without static action–function mappings, and details its implementation and DUET-based evaluation; the codebase is hosted in DUET’s kinesics recognition framework repository \mbox{\cite{duetgithub2}}. Section \mbox{\ref{sec:Discussion}} discusses key findings and implications for closing research gaps, and Section \mbox{\ref{sec:future_work}} concludes with directions for future work.

\vspace{-0.2cm} \section{Taxonomy of Kinesics} \label{sec:Taxonomy} \vspace{-0.2cm}

Humans communicate through a wide range of channels, including spoken language, facial expressions, and body language \cite{HESS2023647}. Among these, body language plays an important role by conveying nonverbal cues that reveal more than the speaker consciously intends. In fact, 93\% of humans' emotional communication relies on body language {\mbox{\cite{Denham2013}}}. These cues are often expressed spontaneously, making them powerful indicators of underlying thoughts and emotions. To study such signals systematically, it is essential to adopt a structured framework, as the diversity of human behaviors makes it infeasible to define a one-to-one mapping between physical movements and their meanings.  For this reason, we draw on the kinesic taxonomy developed by Ekman and Friesen \cite{ekman1969repertoire}. This psychological framework classifies bodily movements according to their communicative function, offering a principled way to interpret the purpose behind specific gestures---particularly those that may not be apparent through verbal communication alone. The taxonomy consists of five categories: emblems, illustrators, adaptors, regulators, and affect displays. 

\vspace{-0.1cm} \subsection{Emblems} \vspace{-0.1cm} 

Emblems are gestures that convey precise, culturally learned meanings \cite{hartman_nonverbal_nodate}. They can supplement verbal communication by repeating, replacing, or adding commentary to spoken messages. By introducing a second layer of information, emblems often enrich interpersonal exchanges. Importantly, the meaning of an emblem is not always universal but socially constructed, varying across cultural contexts. For example, a ``thumbs up'' gesture is interpreted as a sign of approval in many Western cultures but may carry offensive connotations in some Middle Eastern regions. Although emblems are rooted in specific cultural norms, they are not fixed; gestures may migrate across cultural boundaries as individuals observe and adopt them from other contexts. Such multicultural emblems reflect the diverse cultural experiences of users and illustrate the dynamic nature of nonverbal communication.

\vspace{-0.1cm} \subsection{Illustrators} \vspace{-0.1cm} 

Illustrators are gestures that visually reinforce or clarify the spoken message with which they are paired \cite{chute20234}. They are tightly coupled with speech and are often acquired alongside spoken language. A common example is ``pointing,'' which helps direct the listener's attention to a specific object or referent. Beyond providing deictic support, illustrators also reveal the speaker's mental and emotional engagement. A lack of illustrative gestures may indicate fatigue, disinterest, or deliberate verbal control, while their increased use typically signals heightened involvement or enthusiasm. As such, illustrators function as both semantic enhancers and affective indicators within communication.

\vspace{-0.1cm} \subsection{Adaptors} \vspace{-0.1cm} 

Adaptors are self-directed behaviors in which one part of the body manipulates another, such as stroking, scratching, or pressing \cite{neff2011don}. Unlike emblems or illustrators, these movements are typically subconscious and serve no communicative or instrumental goal. Instead, they are primarily used for self-regulation and comfort, often emerging in moments of stress, anxiety, or discomfort. For example, ``scratching the head'' may occur when someone feels nervous or uncertain. Because adaptors operate near the threshold of awareness, they provide a revealing window into a person's emotional state, with their frequency often corresponding to the level of internal tension or discomfort experienced by the speaker.

\vspace{-0.1cm} \subsection{Regulators} \vspace{-0.1cm} 

Regulators are gestures used to manage the flow and rhythm of conversation between multiple interactants. They serve to initiate, maintain, or terminate exchanges between interlocutors, signaling when a speaker should continue, pause, repeat, or yield the floor. For instance, a listener may nod to encourage a speaker to proceed, subtly indicating comprehension and attentiveness. While emblems and illustrators can also influence conversational dynamics, regulators are distinguished by their exclusive role in directing the turn-taking process. They are essential to maintaining conversational coherence and mutual understanding in social interactions.

\vspace{-0.1cm} \subsection{Affect Displays} \vspace{-0.1cm} 

Affect displays are expressive movements that reveal a person's affective and emotional state. For example, crossing one's arms may signal defensiveness, anxiety, or discomfort. Although the meanings of affective expressions tend to be universally recognizable, their usage is shaped by culturally learned ``display rules.'' These rules govern how, when, and to what extent emotions should be inhibited, exaggerated, masked, or substituted in different contexts. Because display rules are influenced by cultural norms and personal upbringing, the visibility and regulation of affect displays vary across individuals and societies. As such, they represent both an innate and socially modulated component of nonverbal communication.

\vspace{-0.1cm} \subsection{Kinesics as a Micro-Interactional Basis for Social Capital} \vspace{-0.1cm} 

Ekman and Friesen's kinesics taxonomy \mbox{\cite{ekman1969repertoire}} provides a principled framework for measuring several specific types of social interactions that build social capital. Social capital theory distinguishes between several fundamental forms based on interaction, including \mbox{\emph{bonding capital}}, which develops through sustained interactions within close relationships characterized by trust and reciprocity \mbox{\cite{putnam_bowling_2000}}, and \mbox{\emph{bridging capital}}, which emerges from weaker ties that span social boundaries and facilitate information flow across networks \mbox{\cite{granovetter1973strength}}. The behavioral vocabulary of kinesics---emblems, illustrators, affect displays, regulators, and adaptors---maps naturally onto these interaction patterns, offering an empirically grounded approach to operationalizing how bodily behaviors build different forms of social connection in future work.

\vspace{-0.1cm} \subsubsection{Bonding Capital: Sustained Exchanges} \vspace{-0.1cm} 

Bonding capital formation depends critically on three classes of kinesic behaviors that characterize sustained, trust-building interactions: regulatory behaviors, affect displays, and low adaptor behaviors. These behaviors create the emotional synchrony, authenticity, and comfort necessary for strong relationship development.

Regulators are fundamental to bonding capital formation. Empirical evidence shows that turn-taking self-disclosure promotes relationship formation: Sprecher et al. \mbox{\cite{sprecher2013taking}} found that alternating, reciprocal sharing led to higher liking, closeness, and enjoyment than extended or non-reciprocal disclosure. This finding aligns with meta-analytic evidence that reciprocal self-disclosure promotes social attraction and positive relational outcomes \mbox{\cite{dindia2002self}}. Complementing this work, Stevanovic and Peräkylä \mbox{\cite{stevanovic2015experience}} argue that the sequential organization of turn-taking can provide a scaffold for contingent emotional reciprocity, enabling participants to share experiences more fully.

Affect displays serve as critical signals in bonding capital formation by conveying trustworthiness and authenticity. Drawing on evolutionary and social-exchange theory, Boone and Buck \mbox{\cite{boone2003emotional}} propose that emotional expressivity acts as a marker of trustworthiness: expressive individuals reveal their intentions through affective cues, enabling partners to assess their reliability and cooperative disposition. The role of authenticity in these displays is crucial. In a series of studies---including interactions coded for observable authenticity behaviors---Rossignac-Milon et al. \mbox{\cite{rossignac2024real}} find that partners judged as more authentic elicit stronger relationship-initiation intentions, an effect driven by the experience of shared reality.

Adaptors function as embodied indicators of internal states such as anxiety, tension, or discomfort, making them essential for understanding bonding capital formation. According to Ekman and Friesen, self-directed movements—including fidgeting, self-touch, and other displacement behaviors—increase under psychological stress and diminish when individuals feel relationally secure \mbox{\cite{ekman1969repertoire}}. Their reduction provides a visible cue that an interaction partner experiences safety and ease, conditions that facilitate the mutual vulnerability required for strong ties. This interpretation is consistent with attachment research: individuals higher in attachment security report greater comfort with closeness and lower relational anxiety \mbox{\cite{collins1990adult}}.

\vspace{-0.1cm} \subsubsection{Bridging Capital: Brief, Acknowledgment-Focused Exchanges}

Bridging capital, in contrast to bonding capital, develops through weaker ties characterized by brief, acknowledgment-focused interactions that facilitate information flow across social boundaries. Such lightweight exchanges—waves, nods, brief greetings, or short conversational contacts—are central to the weak-tie processes described by Granovetter, who emphasizes that infrequent, low-investment social encounters are crucial for transmitting novel information across network clusters \mbox{\cite{granovetter1973strength}}. Putnam similarly conceptualizes bridging capital as rooted in ``thin trust'' that spans diverse social cleavages and depends on minimal yet reliable signals of recognition across difference \mbox{\cite{putnam_bowling_2000}}. Within this framework, two kinesic categories are particularly relevant: emblems, which convey culturally shared signals that support rapid recognition and acknowledgment, and illustrators, which enhance communicative clarity in interactions with limited shared background knowledge.

Emblems play a foundational role in bridging interactions because they function as efficient, low-commitment signals that can be understood even in the absence of a shared interpersonal history. Ekman and Friesen’s foundational work describes emblems as discrete, symbolic acts that enable communication across disparate social groups without requiring extended engagement \mbox{\cite{ekman1969repertoire}}. These gestures serve as boundary-crossing tools: they acknowledge presence, open channels for contact, and reduce social distance in settings where individuals may differ in background, status, or group identity. Their clarity and conventionality make them ideal for sustaining the fleeting, low-stakes interactions that give rise to bridging capital.

Brief illustrators serve a complementary function by facilitating communication across lines of difference. Illustrators---gestures that accompany and augment speech---enhance comprehension and reduce ambiguity, particularly in interactions where participants lack extensive shared context. Research by Kendon \mbox{\cite{kendon2004gesture}} and McNeill \mbox{\cite{mcneill1992hand}} demonstrates that illustrators help interlocutors convey meaning more effectively in first-time or cross-group interactions, increasing communicative efficiency when common ground is thin. Nonverbal immediacy research similarly shows that open hand movements, nodding, and other supportive gestures reduce uncertainty and promote approachability among strangers \mbox{\cite{burgoon2016nonverbal}}.

\vspace{-0.2cm} \section{Dyadic Human Action Recognition} \label{sec:HAR} \vspace{-0.2cm}

HAR is a field within artificial intelligence focused on identifying and analyzing human actions from sensor data, and it has achieved significant success across various domains. The success of HAR can be attributed to many factors, including the commitment of the field to producing publicly available datasets that can be used to help refine data-driven deep learning algorithms across various contexts. While there is an abundance of HAR datasets already available, the majority pertain to single-person---or \textit{monadic}---activities. A better understanding of two-person---or \textit{dyadic}---interactions is essential for enhancing the accuracy, responsiveness, and overall capabilities of systems where human interaction plays a central role. 

Dyadic interactions, which involve the interplay between two individuals, convey deeper communicative and cultural significance. Despite their complexity, HAR for dyadic interactions offers several advantages. The inclusion of a second subject improves the performance of many HAR tasks by introducing an additional distinguishing factor \cite{adeli2020socially}. For instance, consider the actions ``waving in'' and ``thumbs up.'' These two movements appear similar at first glance, as both involve extending one’s arm. However, their small-scale hand movements differ only slightly, making them difficult to distinguish in isolation. The distinction becomes clearer when another subject is involved---specifically, by observing the initiating action of one subject and the reaction of the other. The reacting subject may physically approach if the initiating subject waves them in, while they may simply nod in acknowledgment if the initiating subject gives a ``thumbs up.'' These differing responses provide valuable contextual cues that improve the accuracy of recognizing and differentiating between the two activities. The study of dyadic interactions allows for a more accurate understanding of human behaviors that are absent in monadic activities, enabling systems to better interpret and respond to social dynamics. For example, social robots designed to provide companionship for children leverage dyadic analysis to recognize dangerous situations and intervene in a timely, adaptive manner. These robots also utilize two-person datasets to deliver more natural and engaging conversational interactions, supporting the social, cognitive, and emotional development \cite{chen2022dyadic}. Additionally, in public infrastructure, accurately recognizing dyadic social activities enhances safety by detecting potential dangers and enables the provision of more personalized services in public spaces \cite{coppola2020social}.

\begin{table*}[tb]
  \begin{center}
  \caption{A comparison of existing dyadic datasets shows that our proposed dataset has the \textit{highest number of samples per class}, the most views, and a relatively high number of locations. Note: (1) ``views'' refers to different sensor orientations, (2) ``background noise'' indicates the presence of random people's movement or cluttered environments, and (3) ``D'' is depth, ``J'' is joints. \label{table:dataset_comparison}}
    {\footnotesize{
\begin{tabular}{llc@{\hspace{0.2cm}}c@{\hspace{0.2cm}}c@{\hspace{0.2cm}}c@{\hspace{0.2cm}}cc}
\toprule
\textbf{Dataset} & \textbf{Modalities} & \textbf{\#Videos} & \textbf{\#Classes} & \textbf{\makecell{\#Loca-\\tions}} & \textbf{\#Views} & \textbf{\makecell{Backgro-\\und noise}} & \textbf{Year}\\
\midrule
\makecell[l]{UT Interaction\\\cite{ryoo2010overview}} & RGB & 160 & 6 & 2 & 1 & Partial & 2010\\
\makecell[l]{SBU Kinect\\\cite{yun2012two}} & RGB+D+J & 300 & 8 & 1 & 1 & No & 2012\\
\makecell[l]{JPL Interaction\\\cite{ryoo2013first}} & RGB & 399 & 7 & 5 & First-person & Partial & 2013\\
\makecell[l]{G3Di\\\cite{bloom2016hierarchical}} & RGB+D+J & 168 & 14 & 1 & 1 & No & 2015\\
\makecell[l]{M$^2$I\\\cite{liu_multi-modal_2018}} & RGB+D+J & 1,760 & 9 & 1 & 2 & No & 2015\\
\makecell[l]{ShakeFive 2\\\cite{van2016spatio}} & RGB+J & 153 & 8 & 1 & 1 & No & 2016\\
\makecell[l]{PKU-MMD\\\cite{liu2017pku}} & RGB+D+J+IR & 4225 & 10 & 1 & 3 & No & 2017\\
\makecell[l]{MMAct\\\cite{kong2019mmact}} & \makecell[l]{RGB+J+\\acceleration+\\orientation+\\Wi-FI+Pressure} & 2162 & 2 & 4 & 4+First-person & Partial & 2019\\
\makecell[l]{NTU RGB+D 120\\\cite{liu2019ntu}} & RGB+D+J+IR & 24,828 & 26 & - & 155 & Partial & 2019\\
\makecell[l]{Air Act2Act\\\cite{ko2021air}} & RGB+D+J & 5,000 & 10 & 2 & 3 & No & 2020\\
\midrule
\textbf{DUET (our dataset)} & \textbf{RGB+D+J+IR} & \textbf{14,400} & \textbf{12} & \textbf{3} & \textbf{360} & \textbf{Partial} & \textbf{2025}\\ 
\bottomrule
\end{tabular}
}}
\end{center}\vspace{-0.35cm}
\end{table*}

\vspace{-0.15cm} \subsection{Review of Existing Datasets} \vspace{-0.1cm} 

Despite these advantages, the availability of dyadic datasets remains limited, particularly in comparison to the abundance of monadic datasets. This scarcity poses an increasing challenge as interactions between humans and technical systems grow more complex. The research community's uneven emphasis on these two activities types is reflected in their differing recognition performance levels. \cite{Lin2024c} showed that monadic algorithms, which have achieved outstanding benchmarking records for single-person activities, do not perform nearly as well (i.e., do not transfer) for dyadic interactions. This highlights the disparity between monadic and dyadic activities, which stems from the greater variety of expressive and cultural signals, as well as the increased complexity of spatial and temporal coordination between two or more subjects. To reconcile this discrepancy and improve dyadic HAR, there is a need for more datasets tailored to dyadic interactions. As highlighted in the IEEE Control Systems Society's report on control for societal-scale challenges, traditional boundaries between humans and technology are blurring, and emerging fields like cyber-physical-human systems (CPHS) face challenges in designing robust interactions between humans and control systems \cite{annaswamy2023control}. One of the central CPHS research challenges identified in this seminal report is characterizing how humans adapt during interactions. Dyadic datasets are critical for developing models that can enhance system adaptability, safety, and trustworthiness in these complex built environments \cite{annaswamy2023control}.

Besides increasing the number, diversity, and quality of dyadic datasets, contextualizing activities has proven effective in improving the performance of HAR tasks \cite{niemann2021context}. Contextualization distills meanings embedded in body language, such as emotional and cultural significance, adding another layer of comprehension to the tracking of bodily movements. For instance, a ``thumbs up'' signifies approval in most Western cultures but represents a profanity in Greece and several Middle Eastern countries. Contextualization enables the interpretation of the cultural significance of gestures, such as recognizing the nuanced meanings of a ``thumbs up.'' In addition to enhancing HAR accuracy, contextualization supports the development of various downstream applications. For example, certain branches of CPHS investigate how humans interact with and benefit from the built environment \cite{doctorarastoo2023exploring,DOCTORARASTOO2024115,doctorarastooASCE}. A critical aspect of this framework is understanding the embedded semantics of human behaviors through bodily movements. This understanding provides deeper insights into system use, improving infrastructure design, maintenance, and operation. Contextualization also paves the way for automating psychological and sociological assessments---such as sociometric tests \cite{moreno_foundations_1941}---that rely on self-reported data. These manual evaluations are labor-intensive and also prone to attribution bias. By integrating contextualization with dyadic HAR, these processes can be automated, extracting user preferences from bodily movements \cite{lin2024your} and addressing these limitations. For instance, contextualization enhances telepresence avatars by capturing nonverbal cues and paralinguistic signals, improving the quality and authenticity of remote communication \cite{ahuja2019react}.

Despite these recognized benefits, the few available dyadic datasets---listed in Table \ref{table:dataset_comparison}---are inadequate for extracting the underlying semantics of bodily movements. While some datasets focus on healthcare activities, others are restricted to tracking bodily movements within specific action categories. No existing dataset selects activity classes using scientifically grounded methods that prioritize semantic cohesion to capture the social embeddings of activities. This lack of structured selection limits the ability to understand functional relationships between actions, hindering models from generalizing effectively to new, unlabeled behaviors and new environments. A dyadic dataset that fully supports contextualization is still absent in the research community.

\vspace{-0.2cm} \section{DUET -- \underline{D}yadic \underline{U}ser \underline{E}ngagemen\underline{T} Dataset} \label{sec:DUET} \vspace{-0.2cm}

To enhance HAR performance for dyadic activities through contextualization, we introduce DUET.  Featuring 12 taxonomized interactions, DUET helps to bridge monadic and dyadic HAR while connecting HAR to other disciplines. It is publicly available under an MIT License at \href{https://huggingface.co/datasets/saluslab/DUET}{``DUET''} \cite{huggingfaceAnonymousUploader1DUETDatasets2}.

Instead of repeating previous approaches that arbitrarily select activity categories, our dataset is built on a psychology-based classification that identifies five core kinesics communication functions: emblems, illustrators, affect displays, regulators, and adaptors. Introduced in Section \ref{sec:Taxonomy}, this taxonomy provides a psychologically grounded framework for integrating HAR with interdisciplinary applications. By capturing interactions from all five categories, DUET addresses critical gaps left by existing datasets. This stands in stark contrast to existing datasets, as shown in Table \ref{table:dataset_comparison}, particularly the largest dyadic dataset to date, NTU RGB+D 120 \cite{liu2019ntu}. While NTU RGB+D 120 includes dyadic interactions, it represents only three of the five categories---illustrators, affect displays, and regulators. This imbalance prevents it from supporting applications that require a comprehensive understanding of the taxonomy. In contrast, DUET incorporates interactions from all five categories, ensuring semantic cohesion and preserving the functional relationships between actions. This design enables HAR to generalize more effectively, recognizing both labeled and unlabeled actions by aligning them with shared traits of existing categories. As a result, DUET facilitates connections between HAR and fields like psychology, sociology, and behavioral sciences. By bridging this gap, DUET stands as a critical step toward advancing both HAR and its wider applications.

The remainder of this section presents a detailed and reproducible account of the data collection process and resulting dataset. This includes the sensing modalities and data format, acquisition procedures and setup, subject descriptions, annotation and data management protocols, and the specifications for cross-location and cross-subject evaluations.

\vspace{-0.1cm} \subsection{Data Labels and Modalities} \label{sec:DataLabels}
A total of 12 dyadic activities were selected based on the psychological taxonomy of nonverbal behavior developed by Ekman and Friesen (1969) \mbox{\cite{ekman1969repertoire}}. Dyadic interactions were chosen because the dyad represents the smallest and most fundamental unit of social interaction; higher-order group behaviors can be understood as networks or compositions of these dyadic exchanges \mbox{\cite{brauner_cambridge_2018}}. These activities represent all five categories within the taxonomy, providing a structured foundation for characterizing the kinesic functions of human behavior. The selected interactions are commonly observed in daily life, such as: ``waving in,'' ``thumbs up,'' ``waving,'' ``pointing,'' ``showing measurements,'' ``nodding,'' ``drawing circles in the air,'' ``holding palms out,'' ``scratching hair,'' ``laughing,'' ``arm crossing,'' and ``hugging.'' Their corresponding class labels and kinesic categories are listed in Table \ref{tab:labels}. Because these interactions are prevalent in everyday social encounters, particularly in shared and public spaces, they provide interpretable and transferable representations of human behavior that are directly relevant to modeling sociability, coordination, and responsiveness in built environments and cyber-physical-human systems.

These 12 interactions are collected using the high-quality and multimodal Azure Kinect, equipped with an RGB camera, a depth sensor, and an IR sensor. These sensors all operate at 30 frames per second for three seconds for each video sample, yielding 91 frames per sample. The recorded data is saved in the Matroska (\texttt{.mkv}) container format, allowing multiple tracks of data formats to be extracted through post-processing. Tracks of modalities used in this dataset are RGB, depth, IR, and 3D skeleton joints (or ``keypoints'').

\begin{table}[!t]
    \centering
    \caption{12 interactions selected for the DUET dataset, along with corresponding activity labels and kinesic functions.}
    {\footnotesize{
    \begin{tabular}{ c c c }
    \toprule
    \makecell{Activity label} & Interaction & \makecell{Kinesic function}\\ 
    \midrule
    0 & Waving in & Emblem \\
    1 & Thumbs up & Emblem \\
    2 & Waving & Emblem \\
    3 & Pointing & Illustrator \\
    4 & Showing measurements & Illustrator \\
    5 & Nodding & Regulator \\
    6 & Drawing circles in the air & Regulator \\
    7 & Holding palms out & Regulator \\
    8 & Scratching hair & Adaptor \\
    9 & Laughing & Affect display \\
    10 & Arm crossing & Affect display \\
    11 & Hugging & Affect display \\
    
    \bottomrule
    \end{tabular}}}
    \label{tab:labels}
\end{table}

\begin{figure}[!t]
\centering
\includegraphics[width=0.52\textwidth]{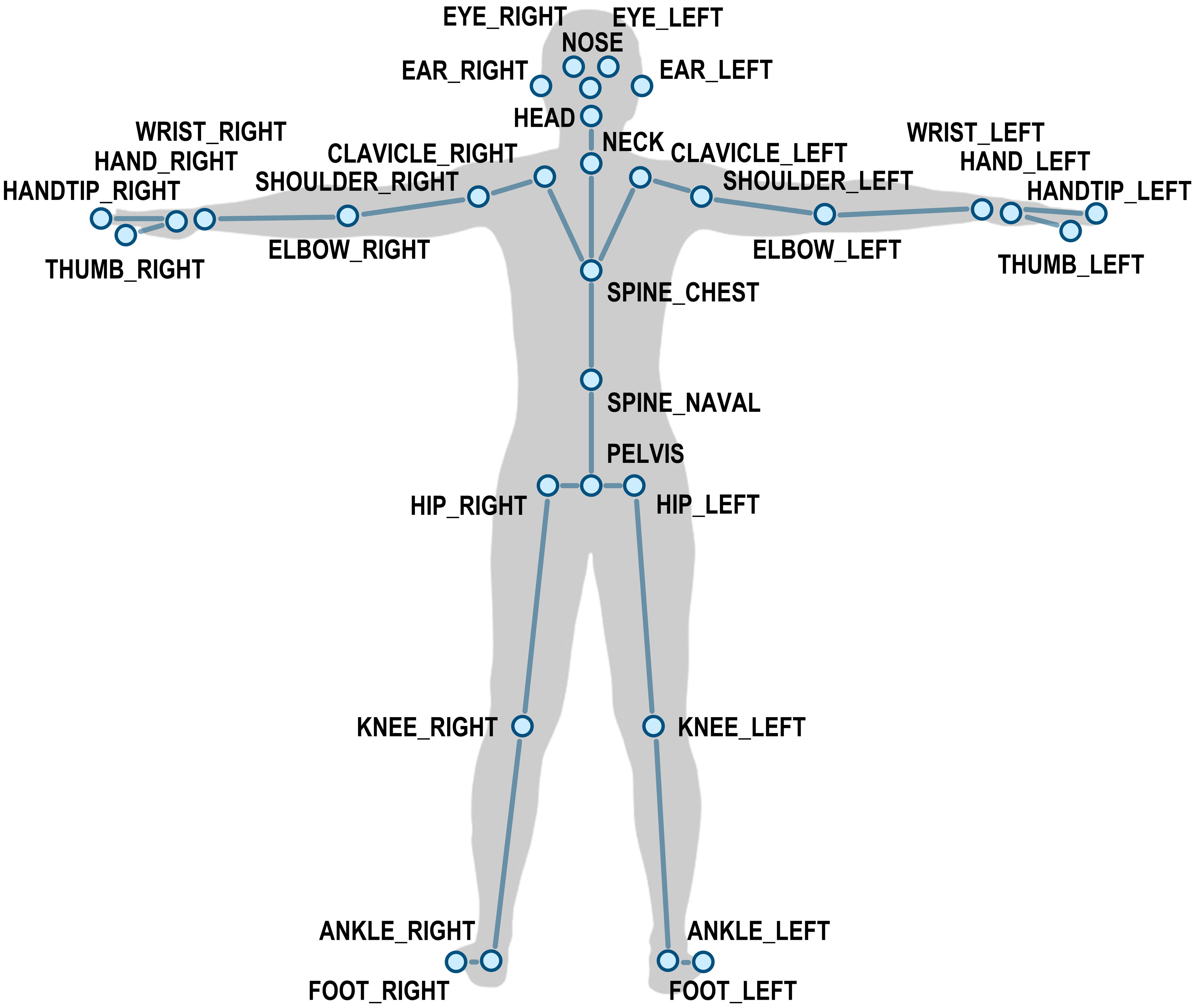}
\caption{Labeling scheme for the 32 skeleton joints extracted using the Azure Kinect \cite{noauthor_azure_nodate}.}
\label{fig:joints} \vspace{-0.45cm}
\end{figure}

The specification of each data format varies depending on the conventions commonly used in the research community: each RGB frame is captured with a resolution of 1,920$\times$1,080 and is stored in a \texttt{.jpeg} format. We record depth and IR sequences with a resolution of 640$\times$576 and store them as 24-bit \texttt{.png} files.  The skeleton joints of every sample video are stored in their corresponding \texttt{.csv} files. Each file contains a 91$\times$193 array, where each row represents a frame, and each column holds information related to that frame. The first column records the timestamp of the frame, and the following 96 columns capture the $x$, $y$, and $z$ coordinates of 32 joints of one subject (as illustrated in Figure \ref{fig:joints}), measured as the distance (in millimeters) from the joint to the camera. For instance, the first three columns record the $x$, $y$, and $z$ values of the first joint. The order of the joints follows the joint index in \cite{noauthor_skeleton_nodate}. The last 96 columns record the 32 joints of the other object.

\begin{figure*}[!t]
\centering
\includegraphics[width=1\textwidth]{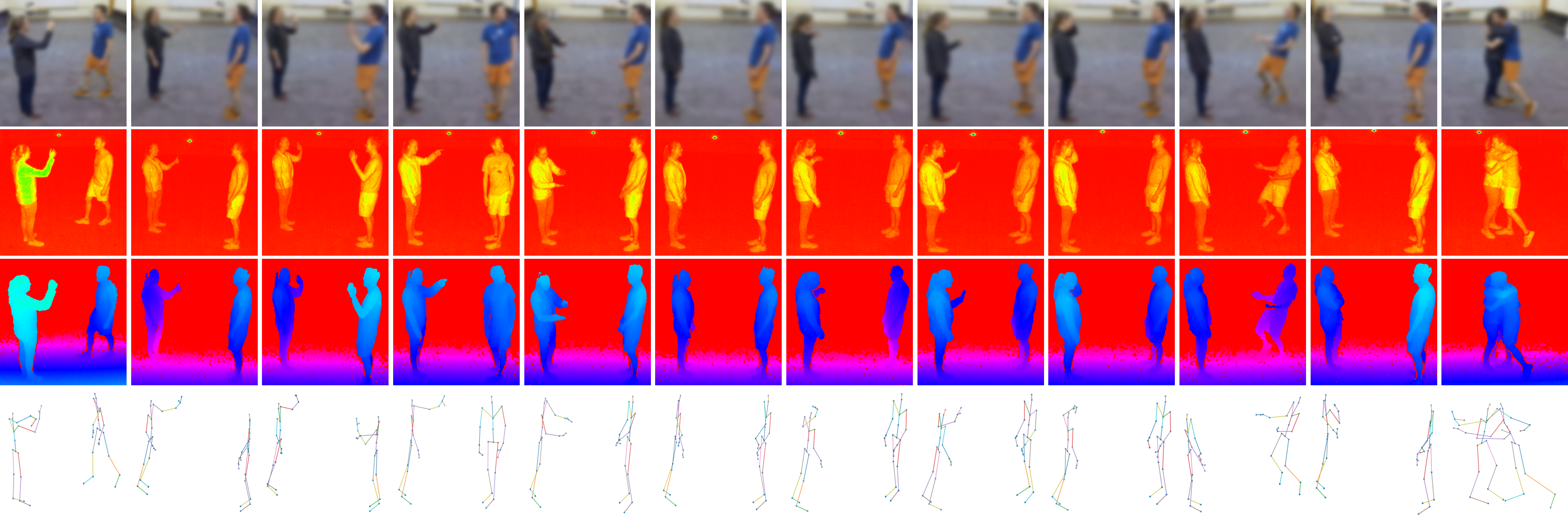}
\caption{Sample data from 12 interactions. The modalities are (top row to bottom row): RGB, IR, depth, and 3D skeleton joints. The 12 interactions are (left to right): ``waving in,'' ``thumbs up,'' ``waving,'' ``pointing,'' ``showing measurements,'' ``nodding,'' ``drawing circles in the air,'' ``holding one's palms out,'' ``twirling or scratching hair,'' ``laughing,'' ``arm crossing,'' and ``hugging.''}
\label{fig:sample_frames} \vspace{-0.45cm} 
\end{figure*} 

Figure \ref{fig:sample_frames} presents sample frames from each action category across different modalities, each offering distinct strengths and weaknesses. RGB frames capture rich details such as interaction types and location, making them highly informative. However, there are two qualities of this modality that detract from its usability in human-centered spaces. First, RGB modality encroaches on user privacy because it captures characteristic features of subjects that reveal their identities. Privacy protection is essential for human-centered applications, such as smart homes, CPHS, and elder care management. Also, RGB often suffers from issues like occlusion and view variation since it compresses the 3D world into a 2D plane. 

In contrast, 3D skeleton joints are less susceptible to these issues. 3D skeleton joints render human bodies in depth images into a number of 3D keypoints that represent the $x$, $y$, and $z$ coordinates of human joints in the 3D world. These keypoints are connected by edges that resemble bones in the human structure. Not only does the skeleton representation remove all characteristic features, it is able to construct the full skeleton even if the subject is partially occluded---the occluded body parts are assigned lower confidence.

In between RGB and 3D skeleton joints are IR and depth sensing, which offer different degrees of privacy and information as illustrated in Figure \ref{fig:sample_frames}. This comparison highlights an inverse relationship between privacy and the value of information conveyed---the more information a modality provides, the less it typically protects user privacy. 

Our dataset includes four modalities that span this entire spectrum, encouraging both the exploration of individual modalities and the fusion of multiple modalities to balance privacy preservation with information richness.

\vspace{-0.25cm} \subsection{Data Acquisition and Setup} 
\begin{figure}[!t]
  \centering
  \begin{subfigure}[b]{0.23\linewidth}
  \centering
    \includegraphics[width=1\linewidth]{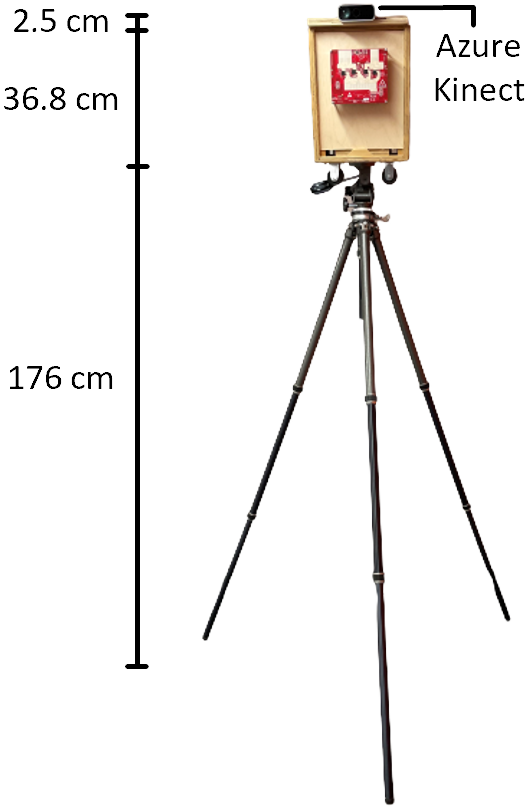} \vspace{-0.2cm}
    \caption{}
    \label{fig:sensing_module}
  \end{subfigure}%
  ~
  \begin{subfigure}[b]{0.27\linewidth}
  \centering
    \includegraphics[width=1\linewidth]{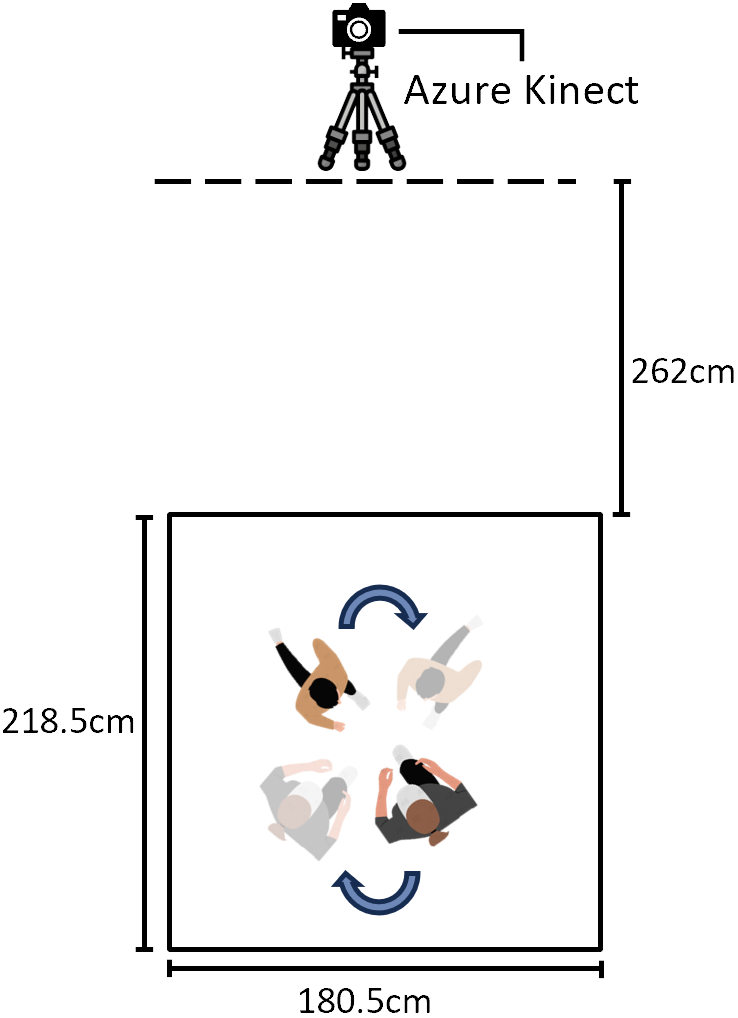}
    \caption{}
    \label{fig:testbed_configuration}
  \end{subfigure}%
  \caption{(a) The sensing module configuration and (b) bird's-eye view of the testbed remain consistent across locations, with subjects confined to a rectangular area.
  }
  \label{fig:setup} \vspace{-0.1cm}
\end{figure}

\vspace{-0.1cm} 
After selecting the Azure Kinect as the sensing device, a custom sensor housing was developed to ensure consistency throughout the experiment. For that reason, we constructed a sensing module, shown in Figure \ref{fig:sensing_module}. The setup places the Azure Kinect 215 cm above the ground and tilts it forward at a 37$^{\circ}$ angle, which captures interactions with a full field of view and minimizes occlusions.

\begin{figure*}[!t]
  \centering
  \begin{subfigure}[b]{0.33\linewidth}
  \centering
    \includegraphics[width=0.95\linewidth, height=3.9cm]{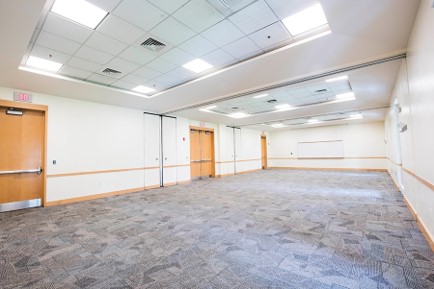}
    \caption{}
    \label{fig:locations-a}
  \end{subfigure}%
  ~
  \begin{subfigure}[b]{0.33\linewidth}
  \centering
    \includegraphics[width=0.95\linewidth, height=3.9cm]{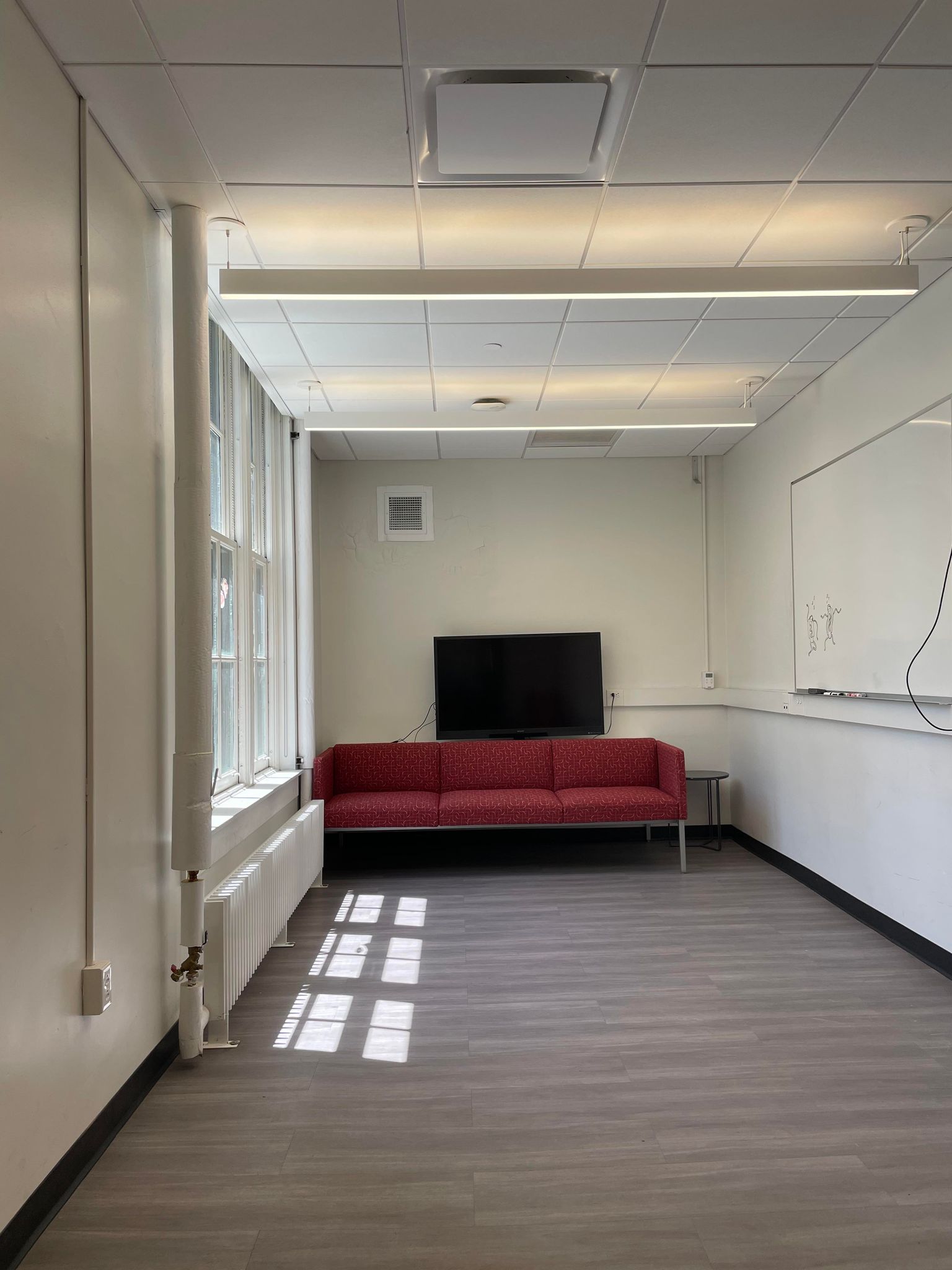}
    \caption{}
    \label{fig:locations-b}
  \end{subfigure}%
  ~
  \begin{subfigure}[b]{0.33\linewidth}
  \centering
    \includegraphics[width=0.95\linewidth, height=3.9cm]{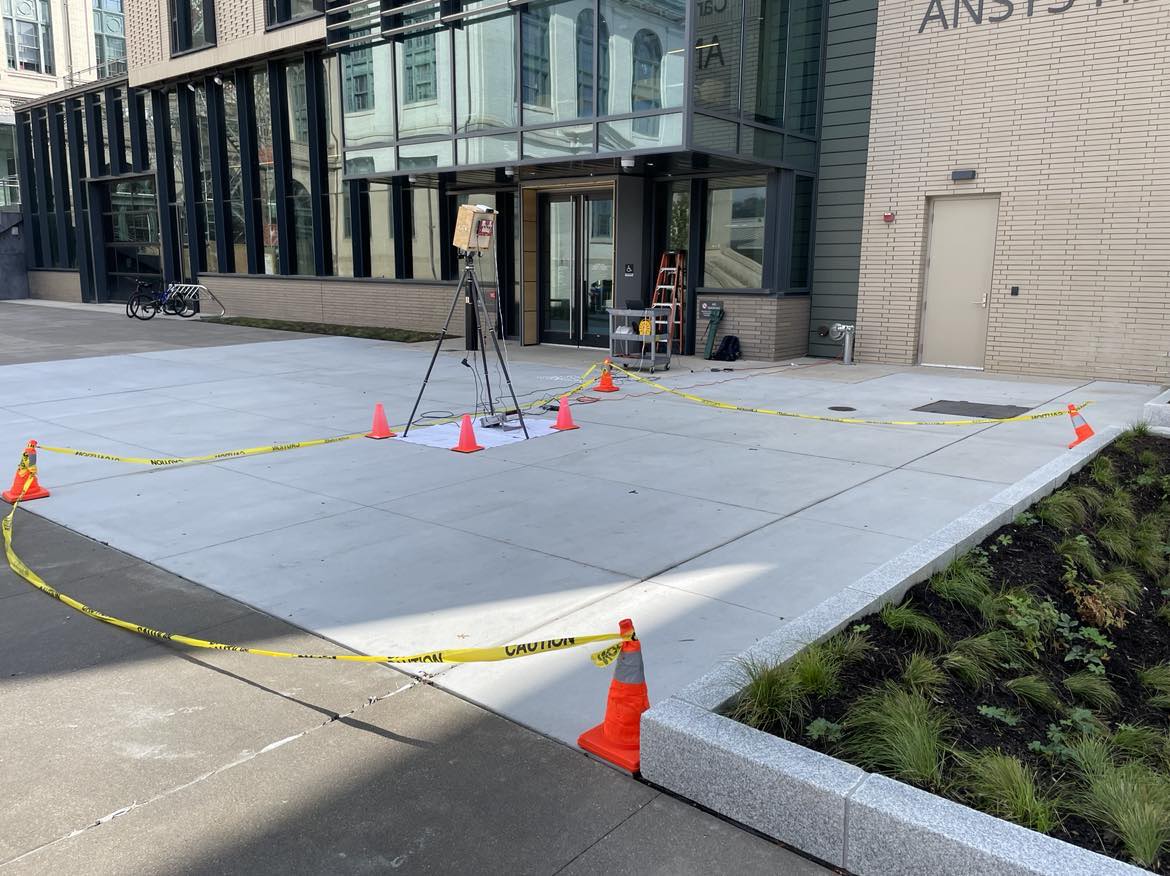}
    \caption{}
    \label{fig:locations-c}
  \end{subfigure}
  \caption{Testing locations: (a) open indoor area, (b) confined indoor space, and (c) outdoor area.
  }
  \label{fig:locations} \vspace{-0.55cm} 
\end{figure*}

An important aspect of the experiment is the selection of testbed locations. The variety of data collection locations provides distinctive backgrounds, minimizing the effect of HAR models overfitting to background noise. As with the number of possible interactions, the number of potential data collection locations is effectively limitless. Instead of attempting to cover all possible environments, we chose three representative locations across Carnegie Mellon University (Pittsburgh, PA): an open indoor area, a confined indoor space, and an outdoor area, as shown in Figure \ref{fig:locations}. These locations typify a wider range of environments where DUET might be applicable, such as parks, schools, nursing facilities, and smart homes. 

\begin{figure*}[!t]
  \centering
  \begin{subfigure}[b]{0.5\linewidth}
  \centering
    \includegraphics[width=1\linewidth, height=12cm]{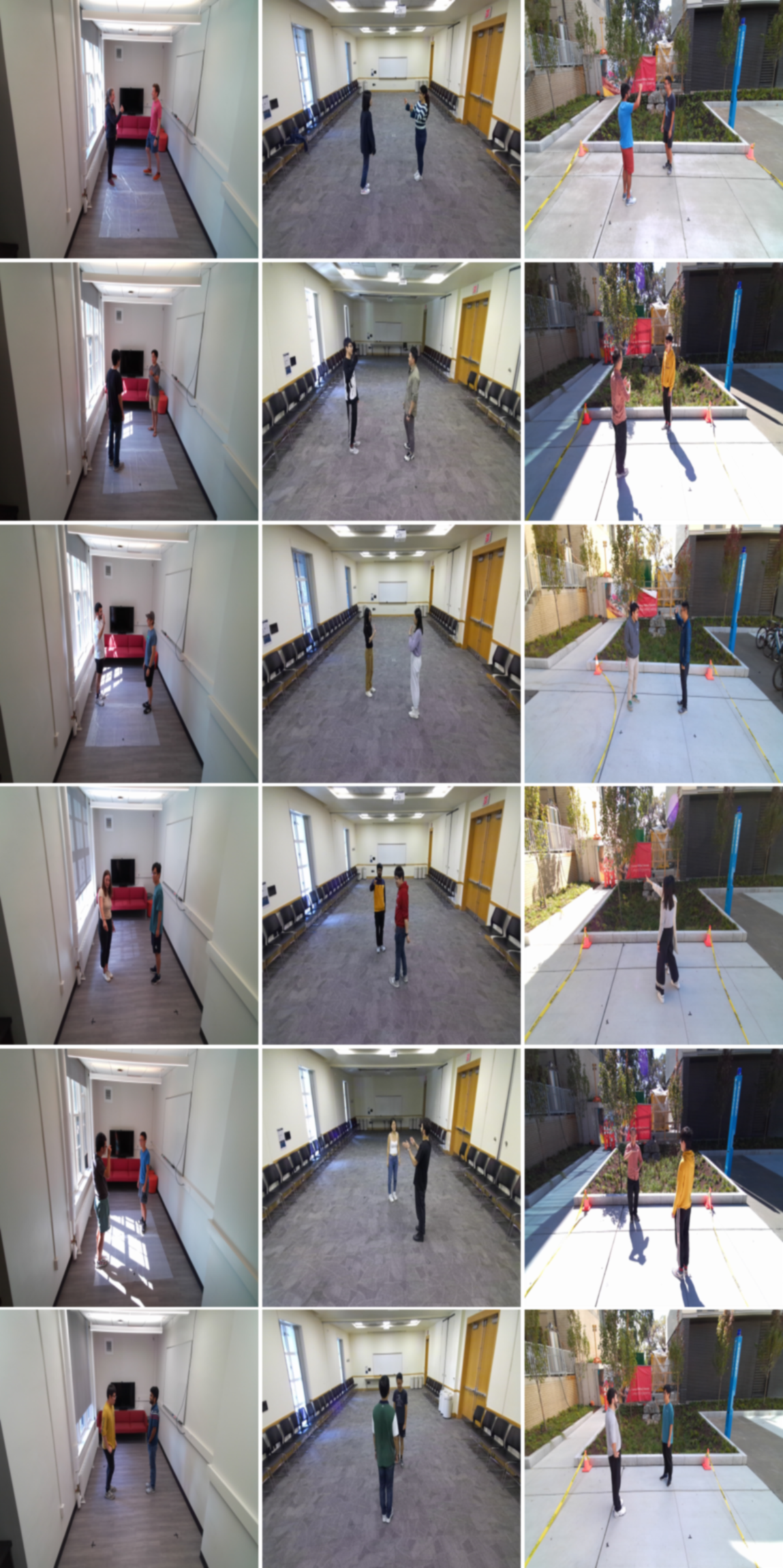}
    \caption{}
    \label{fig:samples-a}
  \end{subfigure}%
  ~
  \begin{subfigure}[b]{0.5\linewidth}
  \centering
    \includegraphics[width=1\linewidth, height=12cm]{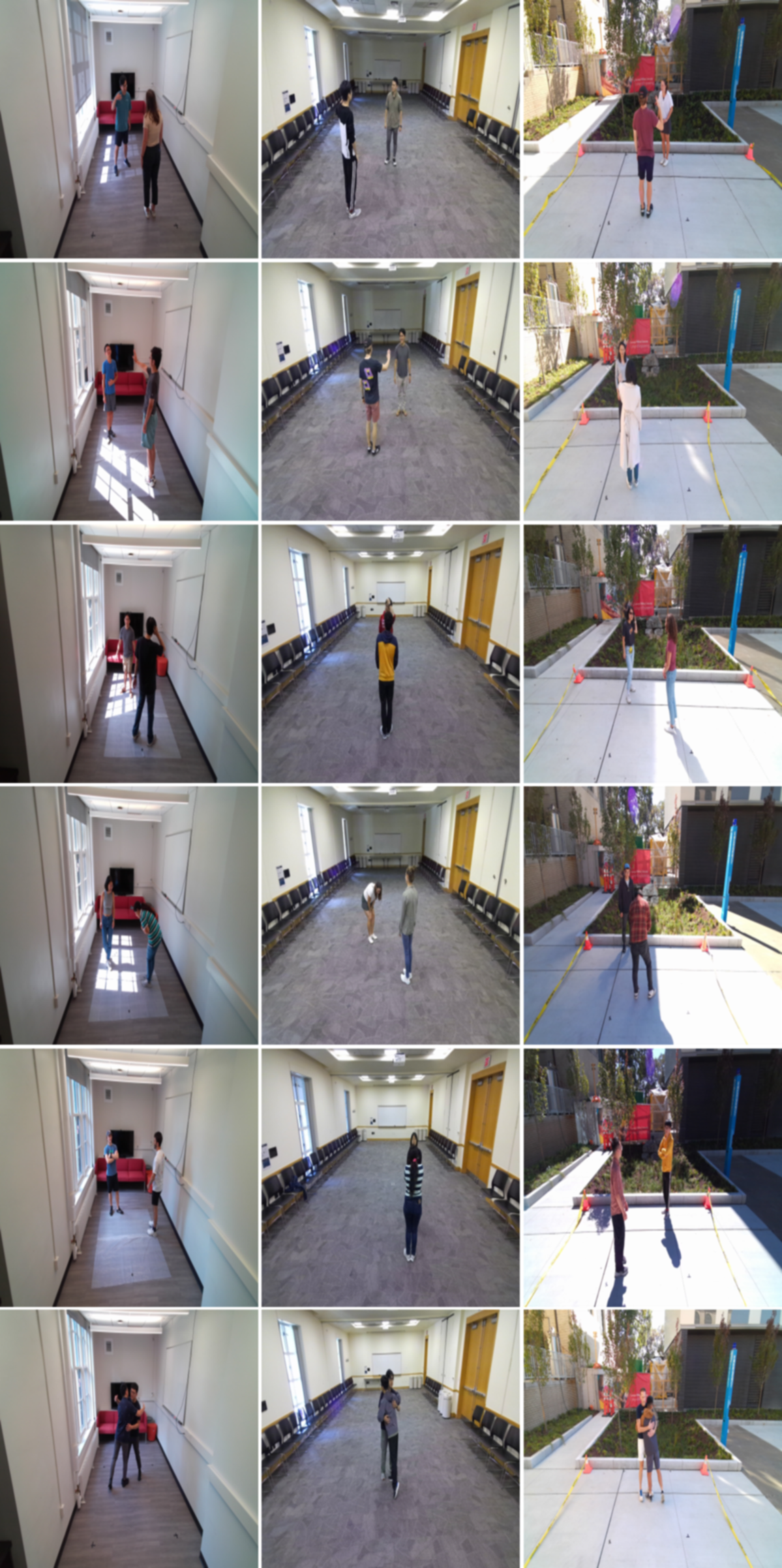}
    \caption{}
    \label{fig:samples-b}
  \end{subfigure}
  \caption{Sample data from the twelve interactions demonstrate the variation in lighting conditions, viewpoints, and occlusions. The locations presented are (left to right): the confined indoor space, the open indoor space, and the open outdoor space. The six interactions are (top to bottom): (a) ``waving in,'' ``thumbs up,'' ``waving,'' ``pointing,'' ``showing measurements,'' and ``nodding,'' and (b) ``drawing circles in the air,'' ``holding one's palms out,'' ``twirling or scratching hair,'' ``laughing,'' ``arm crossing,'' and ``hugging.''
  }
  \label{fig:samples_1} \vspace{-0.45cm} 
\end{figure*}

Furthermore, we intentionally collected DUET data at different times of day and across multiple days to capture a broad range of environmental conditions. For example, in the outdoor setting, some subjects participated during early morning or late afternoon sessions with dim lighting, while others performed at midday under bright sunlight. On several occasions, the sky was fully overcast, providing diffuse low-light conditions. This design choice allowed us to capture diverse illumination scenarios, as illustrated in Figure \mbox{\ref{fig:samples_1}}. In addition to lighting variation, environmental dynamics such as light breezes, strong winds, rustling trees, and an active construction site—whose layout and equipment changed between sessions—further contributed to the heterogeneity of background conditions.

In indoor environments, lighting conditions were varied by adjusting curtain positions to admit different levels of natural light and by reconfiguring overhead lighting for each session. These variations not only enhance DUET’s ecological realism but also mitigate overfitting to static background patterns—an especially important consideration for the RGB modality, which captures both human movement and environmental context. Repeated exposure to an unchanging background can lead machine learning algorithms to erroneously associate background appearance with specific interactions \mbox{\cite{li2023mitigating}}. By contrast, background variation has little to no influence on skeleton-based algorithms, as 3D joint data exclude environmental features and only capture skeleton keypoints within an approximately 6-m range.

Since the experiment was conducted at three different locations, it was essential to ensure the data collection process was consistent and repeatable. We designed a testbed configuration (Figure \ref{fig:testbed_configuration}), which was used across all three environments. Within these 40 repetitions, subjects were encouraged to enact each interaction type following their own interpretations, and the 40 repetitions for the same prompt can be different. For instance, the prompt ``showing measurements'' can be presented in three ways: holding one's arm out with the palm facing down, holding two fingers as if one is pinching a small object, and holding two hands parallel with some distance between them. This flexibility allows us to increase the variation of performance in our dataset. After each repetition, a beep signaled the participants to rotate either clockwise or counterclockwise before proceeding to the next repetition. After the cycle for an interaction was completed, there was a two-minute break before the next interaction started. This structured process helped minimize labeling ambiguity by ensuring that subjects performed each action in a predefined sequence, one action at a time. As a result, we could reliably associate specific images with their corresponding actions, effectively eliminating the potential for ambiguity or labeling errors. In less controlled settings, where actions may overlap or occur simultaneously, we recommend incorporating contextual tags to enhance label clarity and reduce ambiguity in the data. 

Beyond labeling clarity, there are two more benefits for this innovative technique. First, it enabled us to capture interactions from a wide range of orientations relative to the camera. As shown in Figure \ref{fig:samples_1}, some frames capture the side profiles of the subjects, while in others, one subject faces the camera while the other has their back to it. 3D skeleton joints collected with this technique are less sensitive to view variations since their coordinate systems are centered around the same camera with a fixed location. This diversity in orientations enhances the view-invariance of HAR algorithms. Second, our dataset includes samples with occlusions---a common challenge in HAR tasks. Occlusion occurs when one subject fully or partially obstructs the other within the camera's field of view. By incorporating occlusions, our dataset aims to help HAR algorithms address this issue more effectively. Furthermore, capturing multiple viewpoints using a single camera reduces deployment costs, as achieving similar results would otherwise require multiple sensors.

\vspace{-0.2cm} \subsection{Subjects} \vspace{-0.1cm} 

In HAR, an important design consideration is the generalizability of the dataset, which refers to a model’s ability to extract patterns representative of interaction classes in new, unseen performers.  A common practice to improve the generalizability of an HAR dataset—especially in comparison to many prior datasets that include limited demographic spread—is to recruit a sample pool with substantial demographic, physical, and behavioral variation \cite{kong2019mmact,liu2019ntu,ko2021air}.  This ensures that the dataset covers a wide array of presentations of the same interaction class.  To this end, we recruited subjects with distinctly different age, height, weight, and cultural backgrounds.  In terms of physical attributes, a total of 15 male and eight female subjects participated in the experiments.  The subjects were randomly paired to perform actions across the three locations.  The subjects’ ages ranged from 23 to 42 years old, with a mean age of 27 years and standard deviation of 4.01 years.  Heights ranged from 165.1 cm to 185.4 cm, with a mean height of 172.7 cm and a standard deviation of 8.46 cm.  The subjects’ weights ranged from 55 kg to 93 kg, with a mean weight of 69 kg and a standard deviation of 10.1 kg.  These physical differences introduced natural variability in posture, joint positioning, movement amplitude, and timing, all of which are relevant to improving model robustness.

Besides the variance in data contributed by physical attributes, individual and cultural variability in expressing the same interaction class is also accounted for in DUET.  The way that one interprets and performs a gesture can be influenced by cultural norms, upbringing, and situational context, resulting in differences in how body language is manifested and perceived.  Such discrepancies can introduce uncertainty when identifying and classifying kinesic functions, and this challenge extends to automated recognition.  However, this variability also offered an opportunity to evaluate the robustness and generalizability of our framework.  We intentionally recruited participants from diverse cultural backgrounds—including East Asia, South Asia, the Middle East, Europe, North America, and South America—to ensure broad representation of behavioral styles.  Even participants from the same country reported distinct regional or cultural upbringings, which contributed to meaningful within-class diversity in gesture performance.  For example, subjects from several East Asian cultural backgrounds tended to maintain greater interpersonal distance, while others used more expansive gesture ranges.  Despite these differences (and as discussed in Section 6), the trained models achieved strong classification performance across all interaction types, suggesting that the functional categories of kinesics are sufficiently universal to be computationally distinguished even when their physical instantiations vary.

\vspace{-0.2cm} \subsection{Data Format, Annotation, and Management}
\label{sec:data_annotation} 
\begin{figure}[!t]
\centering
\includegraphics[width=0.55\textwidth]{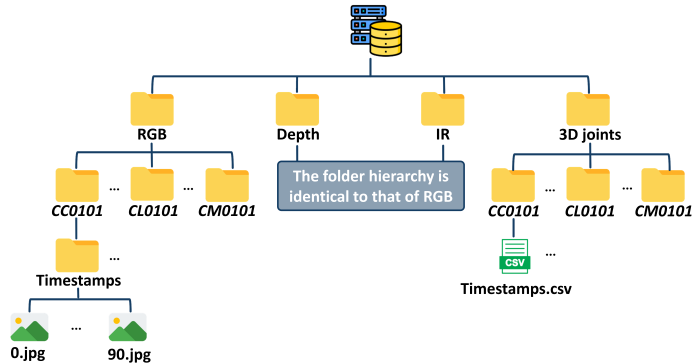}
\caption{The data folder structure for our dataset is designed to ensure easy access for users. The RGB, depth, and IR modalities follow the same hierarchical structure, while the 3D skeleton joint folders store all 3D coordinates for a sample video clip in a single \texttt{.csv} file.}
\label{fig:folder_structure} \vspace{-0.4cm} 
\end{figure}

\vspace{-0.1cm} 
DUET provides four sensing modalities, and---as described in Section \ref{sec:DataLabels}---the specification of each data format varies. To simplify the file compilation, we organized the data into a folder structure, as shown in Figure \ref{fig:folder_structure}. The folder structure comprises four hierarchical layers: (1) modality, (2) location combination, interaction label, and subject, (3) timestamps, and (4) image or \texttt{.csv} files. The first layer classifies files by modality, including RGB, depth, IR, and 3D skeleton joints. The next layer uses a six-digit code, \texttt{LLIISS}, to categorize the location, interaction label, and subject. In this code, \texttt{LL} represents the location: \texttt{CM} for the indoor open space, \texttt{CC} for the indoor confined space, and \texttt{CL} for the outdoor space. \texttt{II} refers to the numbered activities (1--12) listed in Table \ref{tab:labels}, and \texttt{SS} indicates the subject pair, ranging from 1--10. Note that the same subject pair number in different locations does not indicate the same pair; only the pairs \texttt{CCII02} and \texttt{CLII07}, \texttt{CCII01} and \texttt{CMII10}, and \texttt{CCII03} and \texttt{CMII05} represent the same individuals across locations. As mentioned earlier, each pair was asked to repeat an interaction 40 times, and all repetitions were recorded in a single video. To segment the video temporally, we organized each time window by start and end timestamps. For example, a folder named \texttt{40800222\_43800211} contains a recording that begins at 40800222 milliseconds and ends at 43800211 milliseconds after the Azure Kinect is connected. Inside each timestamp folder, the corresponding clip is stored frame by frame, with frames numbered sequentially from 0--90.

\vspace{-0.15cm} \subsection{Cross-location and Cross-subject Evaluations} \vspace{-0.15cm} 
A key motivation for creating DUET is to encourage the research community to explore HAR in the context of dyadic, contextualizable interactions. To support this, we provide a baseline training and test data split for evaluating algorithm performance. In addition to the standard cross-subject evaluation, we also include a cross-location evaluation. We recognize that applications involving dyadic, contextualizable interactions may take place in a variety of indoor and outdoor settings, so the cross-location evaluation helps ensure HAR algorithms are resilient to location variation. For the cross-subject evaluation, we use \texttt{CCII05}, \texttt{CCII07}, \texttt{CLII01}, \texttt{CLII05}, \texttt{CMII06}, and \texttt{CMII09} for the test data, and the remainder for the training data. For cross-location evaluation, \texttt{CCIISS} is selected as the test data, while \texttt{CLIISS} and \texttt{CMIISS} are used as the training data.

\vspace{-0.2cm} \section{Benchmarking State-of-the-art HAR Algorithms} \label{sec:benchmarking} \vspace{-0.2cm}

We evaluate the performance of six open-source, state-of-the-art dyadic HAR models with publicly available code (Table \ref{tab:benchmarking}). We focus specifically on dyadic HAR models because our previous work benchmarked state-of-the-art monadic algorithms on DUET, revealing that monadic models were unable to capture the complex spatio-temporal coordination inherent in dyadic interactions \mbox{\cite{Lin2024c}}. This finding underscores the need for a dedicated dyadic dataset such as DUET, which provides the diversity of classes, samples, and multimodal variation required to advance dyadic interaction recognition. Building on that foundation, the analysis evaluates models that were explicitly developed and trained for dyadic activity recognition, allowing us to assess their capability to capture interactions rich in communicative and psychological function. This work intentionally selects algorithms that are open-source to ensure that the implementation used in benchmarking is consistent with the original benchmarking conducted by the algorithm's developers. This decision prioritizes reproducibility and transparency, both of which are essential for meaningful comparisons. Since DUET provides multiple modalities, the evaluation includes two RGB-based, two depth-based, and two skeleton-based algorithms. The results of the evaluation are in Table \ref{tab:benchmarking}.

\begin{table*}[t]
  \begin{center}
  \caption{Cross-location and cross-subject accuracy comparison for RGB, depth, and 3D skeleton joints. Note: the parenthesized values are accuracies for unoccluded samples. \label{tab:benchmarking}}
    {\footnotesize{
\begin{tabular}{lccc}
\toprule
\textbf{HAR algorithm} & \textbf{Modality} & \textbf{\makecell{Cross-location\\accuracy (\,\%)\,}} & \textbf{\makecell{Cross-subject\\accuracy (\,\%)\,}} \\
\midrule 
\makecell[l]{DB-LSTM \cite{he2021db}} & RGB & 9.65 (13.81) & 17.85 (21.34)\\
\makecell[l]{V4D \cite{zhang2020v4d}} & RGB & 8.26 (18.58) & 7.79 (34.68)\\
\makecell[l]{DOGV-ST3D \cite{xiaopeng2021exploiting}} & Depth & 13.15 & 18.77\\
\makecell[l]{DB-LSTM \cite{he2021db}} & Depth & 14.94 & 23.18\\
\makecell[l]{PAM-STGCN \cite{yang2020pairwise}} & 3D joints & 30.73 & 36.65\\
\makecell[l]{DR-GCN \cite{zhu2021dyadic}} & 3D joints & 38.17 & 41.57\\
\bottomrule
\end{tabular}
}}
\end{center}
\end{table*}

\begin{figure*}[t]
  \vspace{-0.1cm}
  \centering
  \begin{subfigure}[b]{0.33\linewidth}
  \centering
    \includegraphics[width=1\textwidth]{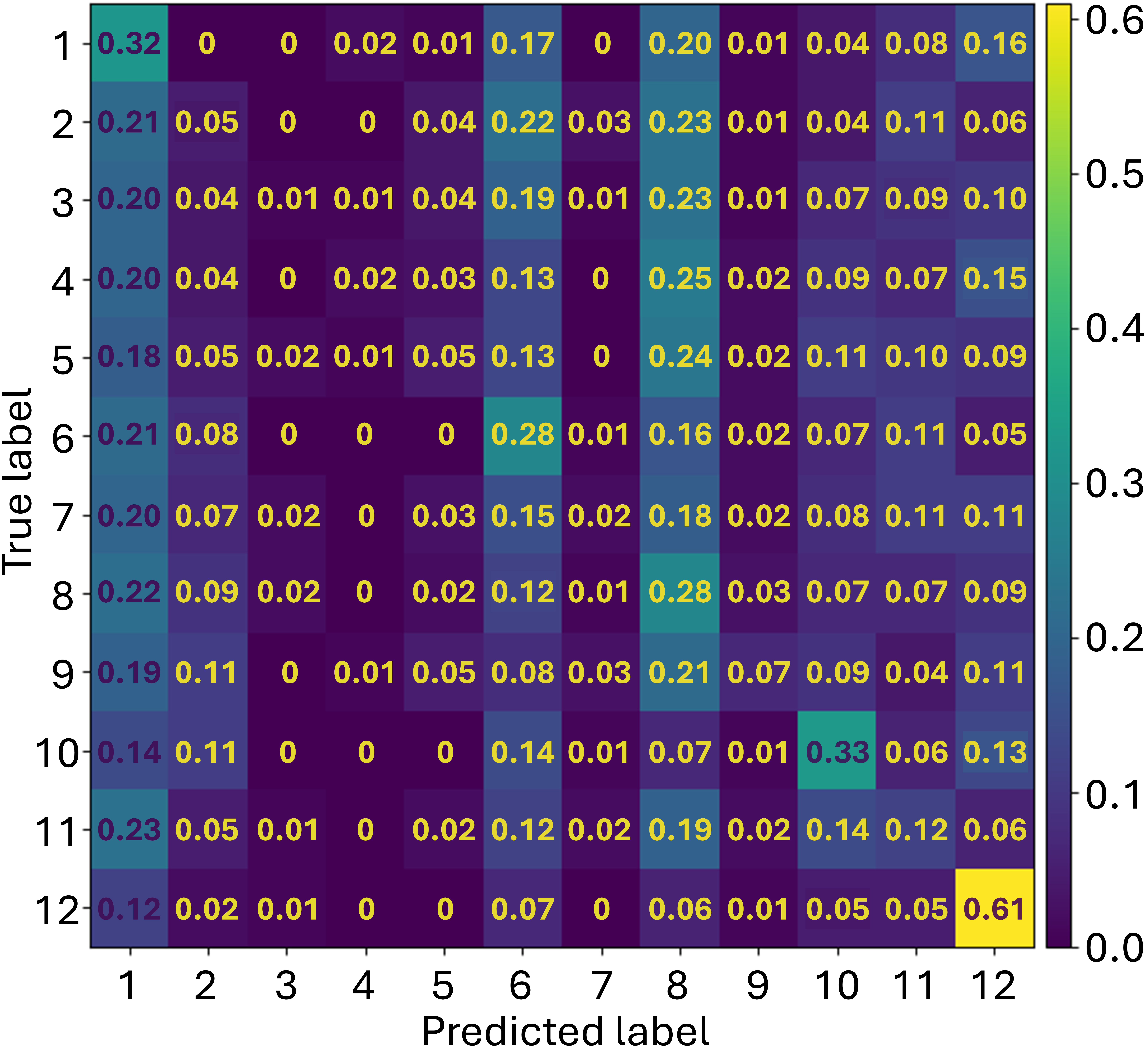}
    \caption{}
    \label{fig:cm-rgb}
  \end{subfigure}%
  ~
  \begin{subfigure}[b]{0.33\linewidth}
  \centering
    \includegraphics[width=1\textwidth]{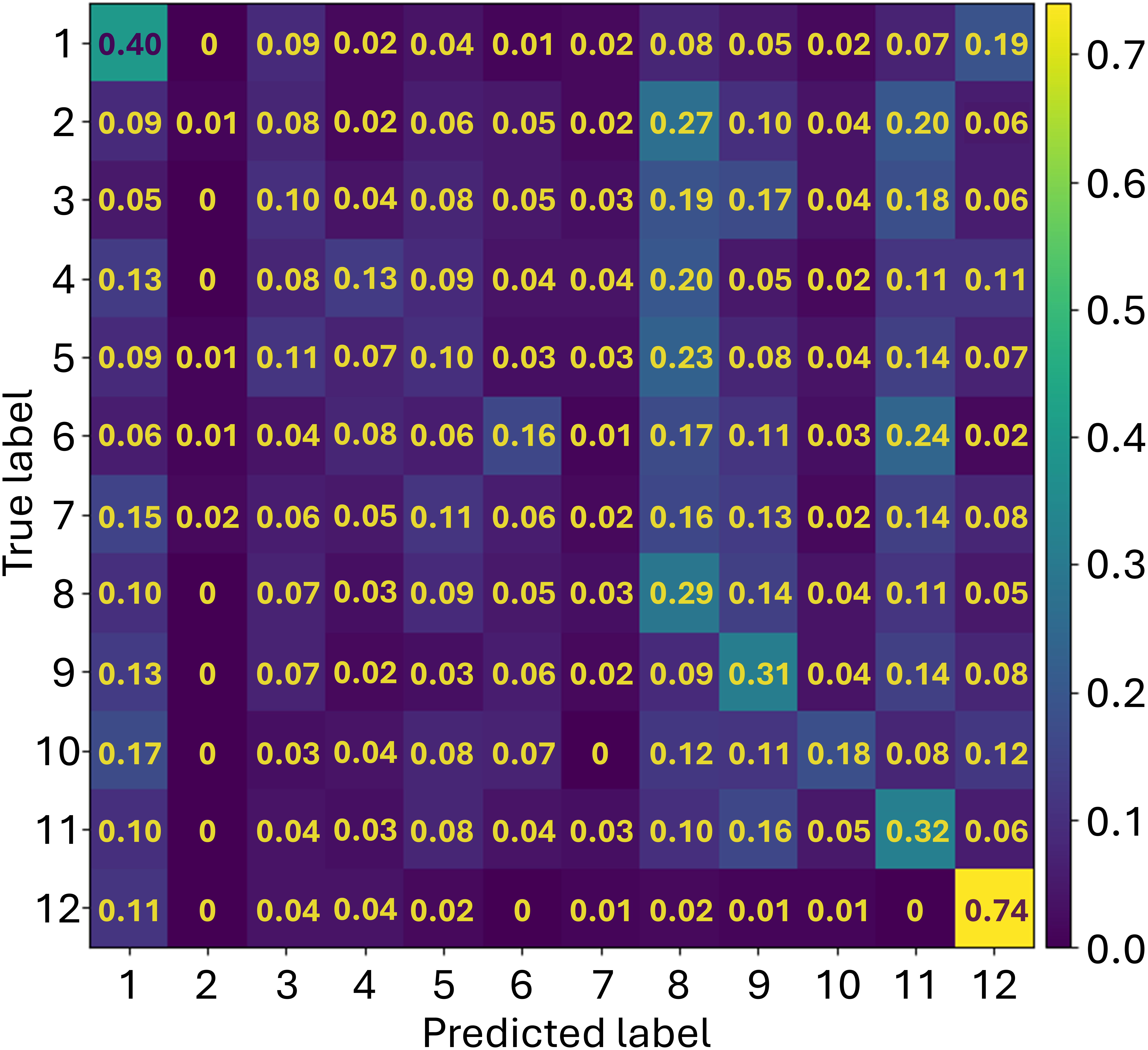}
    \caption{}
    \label{fig:cm-depth}
  \end{subfigure}%
  ~
  \begin{subfigure}[b]{0.33\linewidth}
  \centering
    \includegraphics[width=1\textwidth]{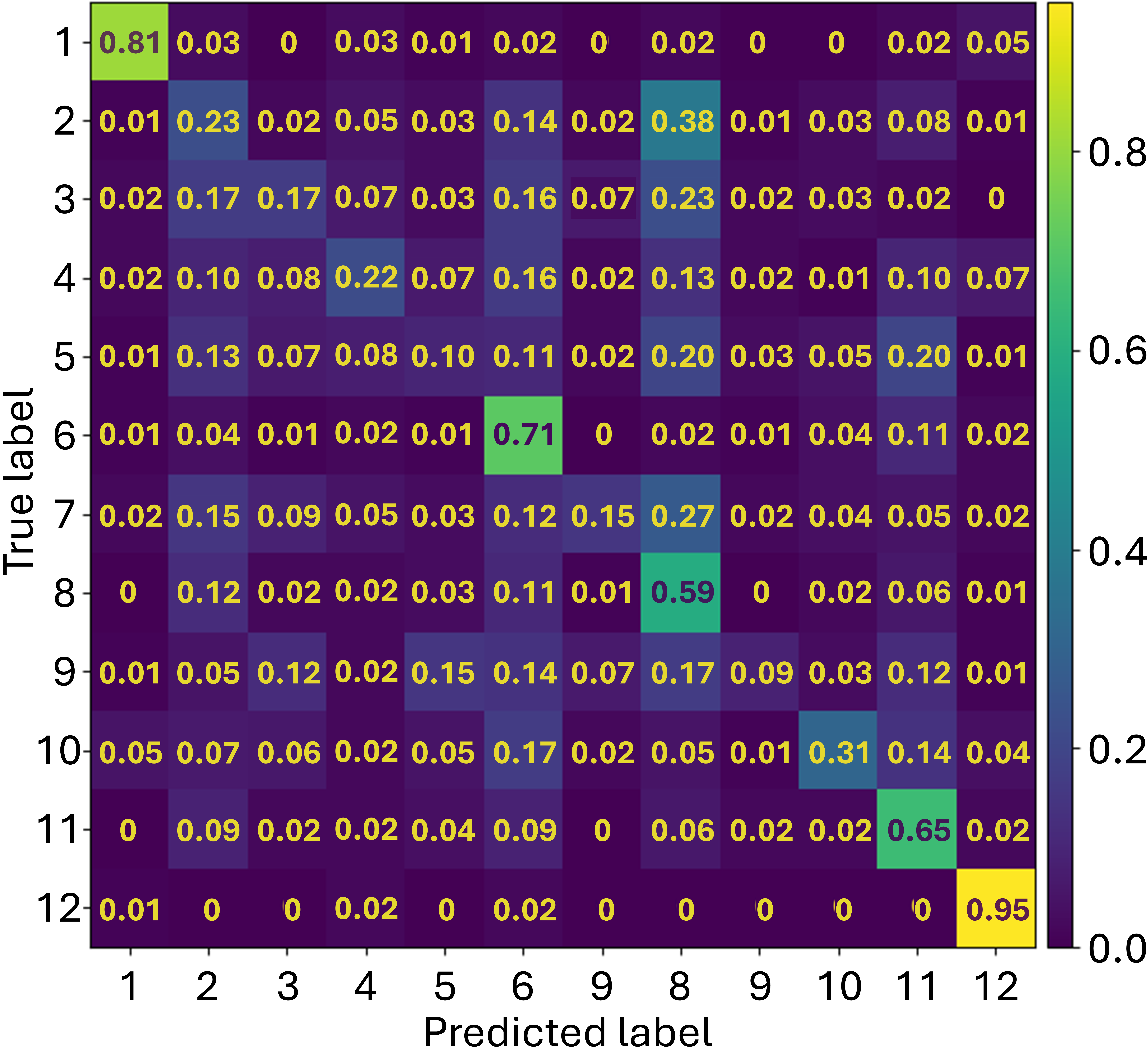}
    \caption{}
    \label{fig:cm-skeleton}
  \end{subfigure}
  \vspace{-0.5cm}
  \caption{Representative confusion matrices for cross-subject evaluation for (a) RGB (DB-LSTM \cite{he2021db}), (b) depth (DB-LSTM \cite{he2021db}), and (c) 3D skeleton joints (DR-GCN \cite{zhu2021dyadic}). Note: each label's interaction corresponds to the mapping in Table \ref{tab:labels}.}
  \label{fig:cm}\vspace{-0.45cm}
\end{figure*}

First, we analyze the effect of occlusion on the RGB modality, as its accuracy is relatively lower compared to other modalities. As previously mentioned, occlusion is a common challenge in vision-based HAR. To evaluate its impact, we train the two selected algorithms using only unoccluded samples for both cross-location and cross-subject evaluations. The corresponding results in Table {\ref{tab:benchmarking}} are comparable to the benchmarking records of other datasets \cite{liu2017pku}. The results indicate that both algorithms show improved performance when occluded samples are excluded. This experiment highlights not only the significant impact of occlusion on algorithm performance but also the critical importance of including occluded samples in datasets for comprehensive evaluation.

Overall, the cross-subject evaluation outperforms the cross-location evaluation across all modalities in the state-of-the-art algorithms, which can be explained by two key factors. First, RGB-based and depth-based algorithms are prone to learning view-dependent motion patterns, often correlating background with motion trajectories during training. In the cross-subject evaluation, the training set includes samples from three locations, whereas in the cross-location evaluation, only two locations are used for training. As a result, these models struggle to generalize to unseen backgrounds during testing, leading to lower accuracy in the cross-location evaluation. Second, the difference in the number of training samples also contributes to the performance gap. In the cross-subject evaluation, 80\% of the dataset (11,520 samples) is used for training, while in the cross-location evaluation, only two-thirds of the dataset is available for training. Performance improves with a larger training sample size. These two phenomena are also present in \cite{liu2019ntu}’s work.

Another observation is the gradual increase in accuracy of the state-of-the-art HAR algorithms tested in our study, progressing from RGB to depth, and then to 3D skeleton joints, which aligns with the expansion of dimensional information. RGB-based algorithms compress input into a 2D plane, leading to lower accuracy since human interactions involve both 3D spatial and temporal coordination \cite{lee2022improving}. This dimensional compression limits the system's ability to fully capture spatial dynamics. Adding depth information to each pixel in an image, as seen in depth-based algorithms, provides an additional layer of information. The improvement in performance with depth inputs is particularly clear when we compare the same model (i.e., DB-LSTM) using RGB and depth inputs separately. However, despite the increase in accuracy from RGB to depth modalities, both still leave room for improvement. This is due to the fact that both modalities operate in Euclidean space (i.e., images), making them more susceptible to view variations. DUET addresses this issue and improves accuracy by providing more robust data. Additionally, training in Euclidean space can be easily influenced by trivial features. As shown in Figure \ref{fig:cm-rgb} and Figure \ref{fig:cm-depth}, RGB and depth models are confused by common poses shared across activities---for example, standing is present in nearly all activities. In contrast, skeleton-based algorithms perform HAR in non-Euclidean space \cite{peng2021rethinking}, representing human interactions in 3D space relative to the camera, leading to better accuracy.

Skeleton-based algorithms outperform other modalities because they capture activities in a 3D space relative to the camera, which is well-suited for the spatial complexity of human interactions. These algorithms can extract underlying motion patterns regardless of the viewpoint. Additionally, 3D skeletons provide a sparse representation of the human body, which helps prevent the network from learning irrelevant features. However, this sparsity can hinder recognition in certain cases. Many dyadic interactions differ only in subtle ways. For example, both the ``thumbs up'' gesture and ``holding one's palms out'' (i.e., activities 2 and 8, respectively) involve arm extension, but the former requires raising the thumb, while the latter involves holding the hand vertically. The simplified skeletal representation may not capture these fine distinctions using current HAR algorithms. This is evident in Figure \ref{fig:cm-skeleton}, which shows these two actions are frequently confused by the algorithm. While the nuances are more apparent in RGB and depth images, from which the 3D skeleton joints are extracted, state-of-the-art skeleton-based algorithms still struggle to detect them.

\vspace{-0.3cm} \section{Embedded Kinesics Recognition Framework} \label{sec:krf} \vspace{-0.2cm}

\begin{figure*}[!t]
\centering
\includegraphics[width=1\textwidth]{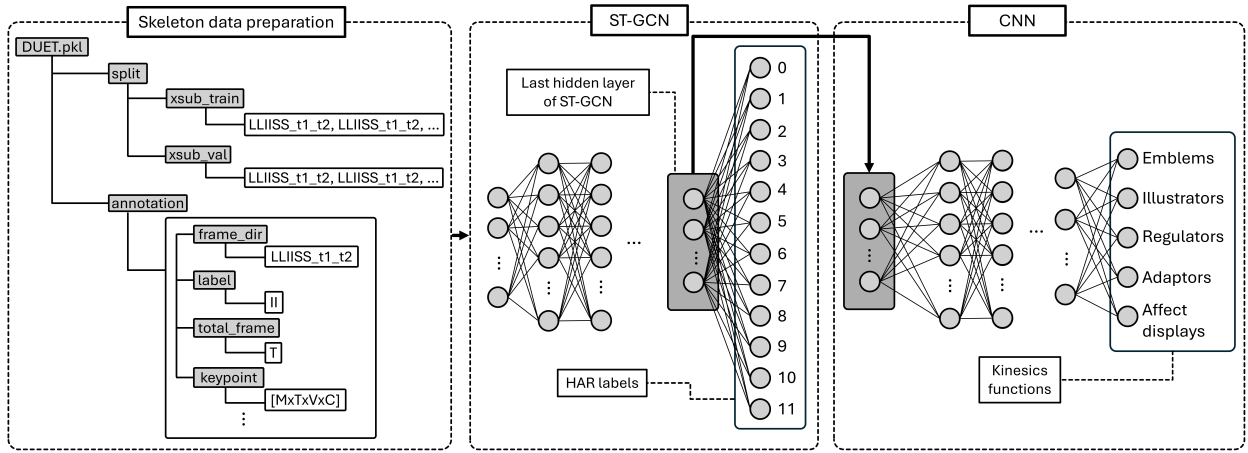}
\caption{Kinesics recognition framework, including data processing, ST-GCN, and CNN. Note: in the data preparation, shaded boxes and the boxes they connect to in the lower layers denote the key-value pairs in the Python dictionary.}
\label{fig:krf_flowchart} \vspace{-0.55cm} 
\end{figure*}

DUET lays the groundwork for understanding the communication clues embedded in bodily movements by integrating a psychological taxonomy with HAR, yet there still lacks a method that infers kinesic functions of human activities from this data. A crude, inefficient, and intractable approach that mirrors existing recognition practices is to append a manually-compiled mapping that translates human activities to their corresponding kinesics to a HAR algorithm. This way, the kinesics of an activity can be determined simultaneously when the activity is recognized. Although this approach successfully identifies the kinesic functions of bodily movements, there is an underlying challenge in the creation of the mapping. Creating a comprehensive mapping that translates all human activities to their kinesics is not feasible since there are countless human activities. In light of this limitation, users would need to tailor the mapping for different applications featuring different activities, which diminishes the generalizability of the framework and escalates the temporal and monetary costs. 

To establish a robust framework applicable for a wide range of applications, we present a framework that distills the kinesic functions from skeleton keypoint data, which is embedded with an intrinsic pattern that clusters different activities with the same communication function together. In addition to illustrating how our approach circumvents the need for an explicit mapping, we carry out a statistical analysis on the framework with DUET to demonstrate its validity. This section includes an overview of the proposed framework, the statistical analysis, and the results. 

\vspace{-0.25cm} \subsection{Kinesics Recognition Framework} \vspace{-0.1cm} 

The kinesics recognition framework is divided into three parts, as demonstrated in Figure \ref{fig:krf_flowchart}, including skeleton data preparation, ST-GCN, and CNN. The data preparation requires converting skeleton keypoint data to comply with the required data format, which is a nested Python dictionary storing the data and its metadata. 

The \texttt{DUET.pkl} file contains two nested dictionaries: \texttt{split} and \texttt{annotation}. The \texttt{split} dictionary specifies the training and validation partitions for ST-GCN and stores them in the lists \texttt{xsub\_train} and \texttt{xsub\_value}, respectively. Every element in the lists is the sample name in the form \texttt{LLIISS\_t1\_t2}. Here, \texttt{II} and \texttt{SS} follow the same notation previously outlined in Section \ref{sec:data_annotation}. Values \texttt{t1} and \texttt{t2} are the start and end timestamps of the interaction. The \texttt{annotation} dictionary stores a list of samples, each containing skeleton data along with associated metadata. Each sample includes the sample name (\texttt{frame\_dir}), activity class label (\texttt{label}), total number of frames (\texttt{total\_frames}), and skeleton keypoints (\texttt{keypoint}). \texttt{frame\_dir} stores the data name, the format of which is identical to the elements in \texttt{split}. \texttt{label} stands for the activity label of the data, which can be a number between 0--11. \texttt{total\_frame} is the number of frames, and \texttt{keypoint} is the skeleton keypoint data array arranged in the shape of \(M\times T\times V\times C\). \(M\), \(T\), \(V\), and \(C\)  stand for the number of subjects (i.e., 2), the number of frames, the number of keypoints (i.e., 25), and dimensions (i.e., $x$, $y$, and $z$ coordinates), respectively.

A notable detail is that we extract only 25 of the 32 skeleton joints provided in DUET, as ST-GCN operates on a reduced set of keypoints. For our experiments, we perform cross-subject evaluation by designating participants \texttt{CCII01} and \texttt{CMII10} as the test set, with the remaining data used for training. This split is applied to both the ST-GCN and CNN components. Once the data are compiled in the required format, they are fed into the ST-GCN model to capture the structural patterns encoded in the skeleton keypoints.

Following data preparation, a transfer learning model comprising ST-GCN and CNN is implemented. The model recognizes kinesic functions of activities through patterns embedded in skeleton keypoint data, which clusters different activities with the same kinesics together. The first requisite for the pattern extraction is ST-GCN. The ST-GCN model consists of 10 layers: one batch normalization layer and nine spatial-temporal graph convolutional operators (i.e., ST-GCN units). The batch normalization layer maintains the scale of all inputs. The nine ST-GCN units have different channels: the first three have 64 output channels, the next three have 128 output channels, and the last three have 256 output channels. Commonly, the output of each ST-GCN unit is randomly dropped out at a 50\% rate to avoid overfitting, and they all share the same temporal kernel size of nine. At the end of the ninth ST-GCN unit, a global pooling was executed on the output tensor to get a 512-dimensional feature vector for each sample, which is then concatenated with a Softmax classifier to determine the sample's activity label. The model adjusts the parameters with stochastic gradient descent. The learning rate is 0.01 and decayed by the rate of 0.1 after every 10 epochs. Typically, ST-GCN is a skeleton-based HAR model that identifies activity types. Instead of using ST-GCN for HAR, the model exploits its pattern recognition capability and encodes keypoint data into lower dimensions in the last hidden layer (i.e., the 512-dimensional feature vector). The learned and condensed pattern in the last hidden layer becomes the input to CNN (Figure \ref{fig:krf_flowchart}). 

The CNN's structure comprises two one-dimensional convolutional layers, the first one has 64 output channels and the second one has 128 output channels. The output from the second convolutional layer is flattened out and fed into a linear transformation layer with the output size of 256 and the drop-out rate of 50\%, consistent with the rationale mentioned earlier. The hidden variables are then fed into a linear transformation of the same output size as the previous layer before determining the kinesics of the associated input activity with a final Softmax layer. The activation function for all the layers, except for the last linear transformation layer, is ReLU. For readers interested in implementing the framework, the codebase for the model and experiments is open-sourced at our kinesics recognition framework repository \cite{duetgithub2}.

\begin{table*}[!t]
    \centering
    \caption{30 subsets of the DUET dataset are tested on the proposed kinesics recognition framework.}
    {\footnotesize{
    \begin{tabular}{ c c c c c }
    \toprule
    \makecell{Experiment\\number} & \makecell{Number of\\interactions} & Activity labels & \makecell{ST-GCN\\Accuracy (\%)} & \makecell{CNN\\Accuracy (\%)}\\ 
    \midrule
    0 & 8 & 0, 1, 3, 4, 5, 8, 9, 10 & 70 & 63 \\  
    1 & 7 & 0, 1, 3, 6, 7, 8, 10 & 31 & 28 \\
    2 & 6 & 0, 1, 3, 6, 8, 10 & 74 & 72 \\  
    3 & 8 & 2, 3, 4, 5, 8, 9, 10, 11 & 78 & 68 \\
    4 & 7 & 0, 2, 3, 5, 7, 8, 11 & 80 & 58 \\
    5 & 6 & 0, 4, 5, 6, 8, 11 & 85 & 74 \\
    6 & 7 & 1, 3, 4, 6, 8, 10, 11 & 34 & 28 \\
    7 & 7 & 0, 2, 3, 4, 7, 8, 11 & 82 & 77 \\
    8 & 5 & 2, 3, 5, 8, 9 & 74 & 55 \\
    9 & 6 & 1, 4, 5, 8, 9, 10 & 57 & 55 \\
    10 & 8 & 0, 2, 3, 6, 7, 8, 10, 11 & 82 & 80 \\
    11 & 7 & 2, 3, 4, 5, 6, 8, 9 & 69 & 67 \\
    12 & 6 & 0, 1, 3, 6, 8, 11 & 61 & 50 \\
    13 & 10 & 0, 1, 2, 3, 4, 6, 8, 9, 10, 11 & 69 & 70 \\
    14 & 11 & 0, 1, 2, 3, 4, 6, 7, 8, 9, 10, 11 & 60 & 53 \\
    15 & 10 & 0, 1, 2, 3, 4, 7, 8, 9, 10, 11 & 58 & 58 \\
    16 & 7 & 1, 2, 3, 5, 8, 10, 11 & 70 & 55 \\
    17 & 11 & 0, 1, 2, 3, 4, 5, 6, 7, 8, 9, 10 & 33 & 39 \\
    18 & 5 & 1, 3, 5, 8, 10 & 73 & 55 \\
    19 & 5 & 1, 4, 6, 8, 10 & 28 & 20 \\
    20 & 11 & 0, 1, 2, 4, 5, 6, 7, 8, 9, 10, 11 & 56 & 51 \\
    21 & 6 & 0, 4, 7, 8, 9, 11 & 66 & 61 \\
    22 & 11 & 0, 2, 3, 4, 5, 6, 7, 8, 9, 10, 11 & 68 & 56 \\
    23 & 10 & 0, 1, 2, 3, 5, 7, 8, 9, 10, 11 & 41 & 39 \\
    24 & 10 & 0, 1, 2, 3, 4, 5, 7, 8, 10, 11 & 56 & 46 \\
    25 & 12 & 0, 1, 2, 3, 4, 5, 6, 7, 8, 9, 10, 11 & 58 & 55 \\
    26 & 8 & 2, 3, 4, 5, 6, 7, 8, 11 & 76 & 62 \\
    27 & 12 & 0, 1, 2, 3, 4, 5, 6, 7, 8, 9, 10, 11 & 55 & 46 \\
    28 & 12 & 0, 1, 2, 3, 4, 5, 6, 7, 8, 9, 10, 11 & 65 & 46 \\
    29 & 5 & 2, 3, 6, 8, 10 & 82 & 79 \\
    \bottomrule
       &  &  & \makecell{\textbf{Pearson correlation} \textbf{coefficient \(\rho\)}} & \textbf{0.91} \\
       &  &  & \makecell{\textbf{95\% Confidence interval of \(\rho\)}} & \textbf{[0.82, 0.96]} \\
    \end{tabular}}}
    \label{tab:results}
\end{table*}

\vspace{-0.15cm} \subsection{Experiment on DUET} \vspace{-0.1cm} 

To evaluate the performance of the proposed kinesics recognition framework, we implemented the framework on DUET. Specifically, we conducted 30 experiments with subsets of DUET consisting of different numbers and types of activities, as shown in Table \ref{tab:results}. This design choice was motivated by the poor performance of ST-GCN on the full DUET dataset, which led to inaccurate low-dimensional representations of skeletal keypoints and, in turn, hindered the CNN’s ability to capture the structural patterns embedded in the keypoint data. Because the two components of the framework play distinct roles—ST-GCN learning a spatiotemporal latent representation, and CNN mapping that representation to communicative function—we report their accuracies separately to evaluate representation quality and function recoverability. To address this issue, we created 29 additional subsets of the DUET dataset and performed statistical analysis using the Pearson correlation coefficient and p-values to examine the relationship between ST-GCN and CNN. In total, 30 subsets were analyzed to ensure the reliability of the correlation estimates \cite{bonett2000sample}. For each subset, the number and types of interactions are randomly selected, as listed in Table \ref{tab:results}. We run the framework with the selected data subsets, where, within all subsets, \texttt{CCII01} and \texttt{CMII10} are used as test data and the rest as training data for both ST-GCN and CNN.

\begin{figure}[t]
  \vspace{-0.15cm} 
  \centering
  \begin{subfigure}[b]{0.4\linewidth}
  \centering
    \includegraphics[width=1\linewidth]{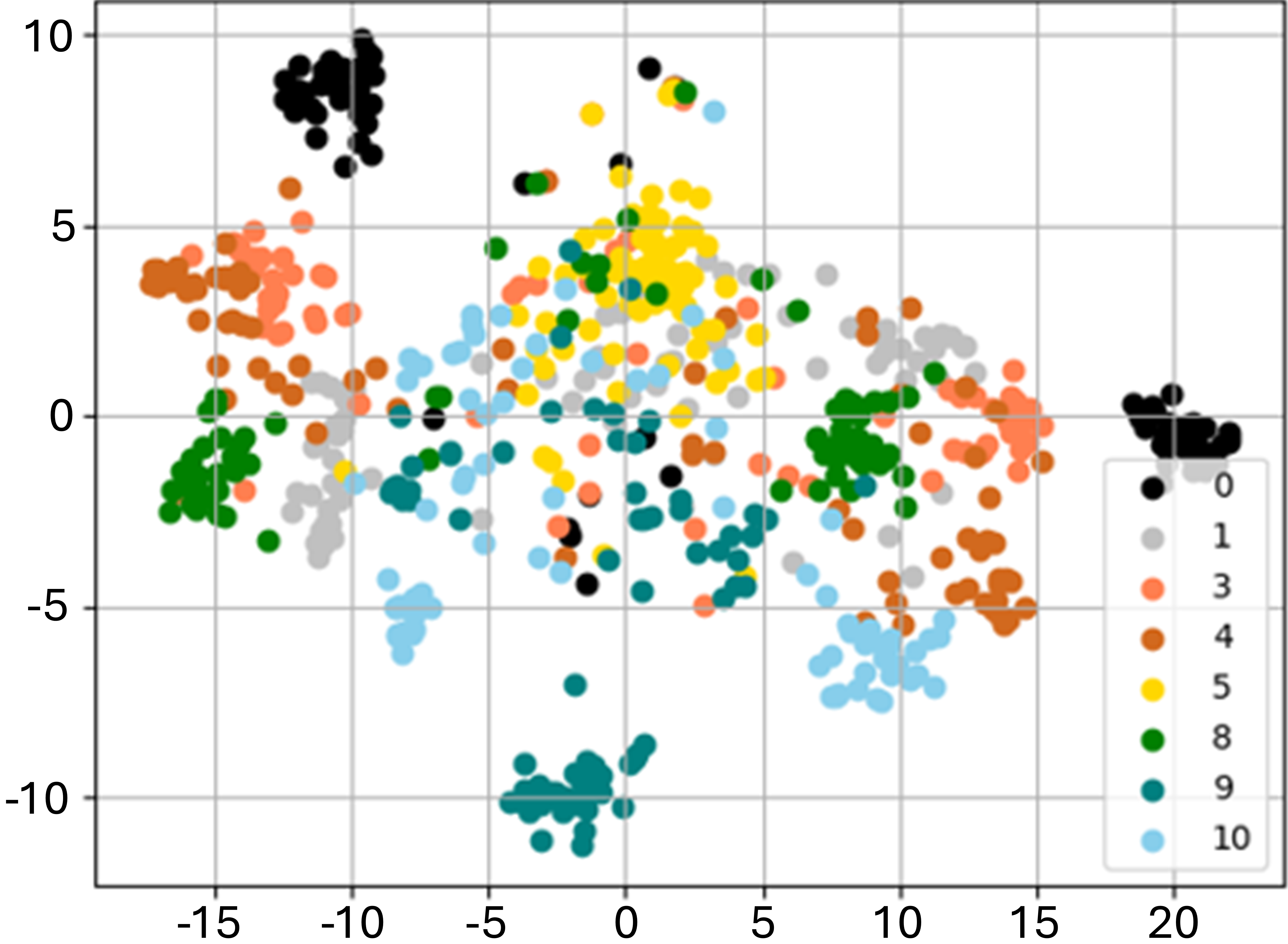}
    \caption{}
    \label{fig:latent_features-a}
  \end{subfigure}%
  ~  
  \begin{subfigure}[b]{0.4\linewidth}
  \centering
    \includegraphics[width=1\linewidth]{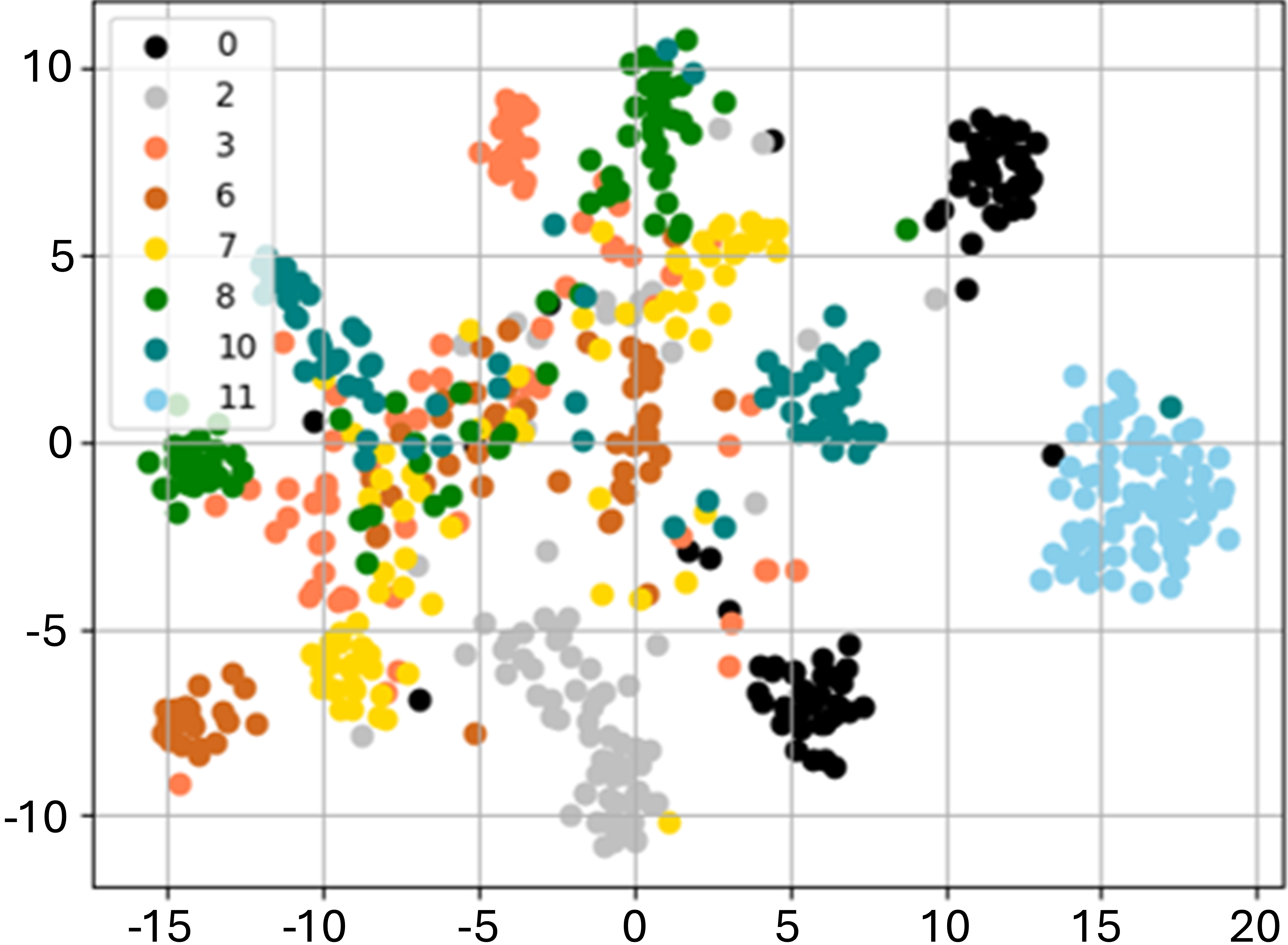}
    \caption{}
    \label{fig:latent_features-b}
  \end{subfigure}

  \caption{The latent features of (a) experiment 0 and (b) experiment 10, which are both reduced to two dimensions with t-SNE dimension reduction technique.
  }
  \label{fig:latent_features} \vspace{-0.6cm} 
\end{figure}

The results in Table \ref{tab:results} show that the performance of the ST-GCN and CNN models improves in parallel---that is, CNN accuracy increases as ST-GCN accuracy increases. This relationship is graphically demonstrated in Figure \ref{fig:latent_features-a} and Figure \ref{fig:latent_features-b}, where the cluster that dictates the kinesics is more apparent in the one with better ST-GCN accuracy (Figure \ref{fig:latent_features-b}). To quantify this relationship, we compute the Pearson correlation coefficient ($\rho$) between the ST-GCN and CNN accuracy values. We use correlation here because our central hypothesis predicts that improvements in latent representation quality should yield improvements in kinesics classification; a strong correlation therefore supports that communicative function is embedded in the learned spatiotemporal structure. Because the experiment does not exhaustively cover all possible interaction combinations, we perform a one-tailed p-value test to assess whether a statistically significant positive linear correlation exists between the two models' accuracies. For this test, we set the significance level ($\alpha$) to 0.05. The null hypothesis (\(H_0\)) and the alternative hypothesis (\(H_a\)) are defined as follows:

\vspace{-0.25cm} \begin{itemize}
  \item \(H_0\): There is no positive linear correlation between ST-GCN and CNN accuracies (\(\rho\leq0\)).
  \item \(H_a\): There is a positive linear correlation between ST-GCN and CNN accuracies (\(\rho>0\)).
\end{itemize}
\vspace{-0.25cm} \noindent Based on the data in Table \ref{tab:results}, the Pearson correlation coefficient, \(\rho\), is 0.91, with an associated p-value of 0.000002. In the case of p-value smaller than \(\alpha\), \(H_0\) is rejected. This result indicates a statistically significant positive linear correlation between the performance of the ST-GCN and CNN models, with a strong correlation coefficient of \(\rho=0.91\). 

This result has two key implications. First, it provides support for the claim that there exists a statistically significant correlation between ST-GCN and CNN accuracies. This means that improving the learned latent representation (as measured by ST-GCN accuracy) directly enhances downstream function recognition, validating the architectural logic of the two-stage framework. Second, since this framework relies on the pattern recognition capabilities of both ST-GCN and CNN, the results suggest the existence of an underlying structure within the HAR data that enables accurate categorization of kinesic functions without relying on predefined mappings. The finding of this structure is especially significant considering our intentional effort to increase the complexity of the dataset by recruiting a physically and culturally diverse subject pool. This emergent structure offers a more efficient and scalable approach to kinesics recognition, reducing both time and cost.

\vspace{-0.2cm} \section{Discussion} \label{sec:Discussion} \vspace{-0.2cm}

In this work, we introduce DUET, a contextualizable dataset comprising 12 dyadic interactions, based on a psychological taxonomy that organizes human interactions into five groups according to their communication functions. Through benchmarking and development of a new kinesics recognition framework, this taxonomy-based work advances HAR beyond simple body movement tracking by extracting the embedded semantics in dyadic interactions. This framework provides a means to measure and quantify the behavioral processes through which social capital, among other human-centered benefits, is produced within the built environment. This offers a measurable and observable basis for understanding the social benefits of infrastructure---benefits that we hope can one day be designed for with the same rigor as infrastructure performance or safety.

While this study did not implement a real-world deployment outside of the creation of the dataset, the proposed framework can support continuous operation (i.e., simulation or design based on near real-time kinesics recognition, if needed). Once skeletal keypoints are streamed from commodity RGB-D sensors, inference using the trained ST-GCN and CNN can proceed, with frame rates of approximately 30 frames per second \mbox{\cite{noauthor_buy_nodate}}. The primary latency arises from preprocessing---particularly keypoint extraction. Even with this computational latency, our framework represents a fundamental shift from the way human states are traditionally inferred in behavioral and environmental studies. Conventional methods such as surveys or ecological momentary assessments capture psychological states at discrete intervals, offering valuable but episodic snapshots of human experience. In contrast, our work supports future deployments necessitating temporally continuous estimation of communicative and psychological functions directly from bodily movement.

To develop DUET, we collected 14,400 samples across 12 interaction types---yielding 1,200 samples per class, the highest sample-to-class ratio among published HAR datasets. The dataset spans four sensing modalities---RGB, depth, IR, and 3D skeleton joints---each offering distinct strengths for modeling behavior. In addition to its scale, DUET is designed to enhance view and background invariance. We introduce a novel data collection procedure that captures interactions from multiple angles using a single camera---an advancement difficult to achieve even with multi-sensor setups. This approach increases robustness to viewpoint variation, reflects real-world observational scenarios, and reduces deployment costs. Data were collected across three distinct environments---an open indoor area, a confined indoor space, and an outdoor location---to promote generalizability and enable the study of environmental context on recognition accuracy.

To establish baseline performance on DUET, we evaluated six open-source HAR algorithms---two each based on RGB, depth, and skeleton data. While some prior work has applied monadic HAR models to dyadic tasks, our results reveal a consistent performance gap. In this study, we extend this analysis by benchmarking six dyadic HAR algorithms on DUET. The findings underscore (1) the underexplored complexity of social interactions and (2) the vulnerability of existing HAR pipelines to environmental and viewpoint variability---both of which open new research avenues.

The benchmarking results presented in Section \ref{sec:benchmarking} serve a dual purpose. On the surface, they offer a rigorous model comparison on a novel dataset. More fundamentally, they highlight the limitations of current HAR algorithms when applied to dyadic interaction, where semantics are more subtle and context-dependent. The relatively poor performance of state-of-the-art models---despite success on monadic benchmarks---suggests that DUET reveals behavioral structure that is not well captured by existing techniques. In this sense, DUET does not merely replicate prior datasets but actively challenges and extends them. DUET provides a testbed that exposes foundational gaps in model generalization, interaction modeling, and semantic inference---validating its value as a critical resource for advancing socially intelligent HAR systems in diverse environments.

Building on this foundation, the second phase of our discussion transitions from identifying limitations to proposing new capabilities. DUET's taxonomy-based structure enables a reconceptualization of the recognition task itself---not simply identifying the visible activity, but inferring the communicative function of that activity. This shift---from action recognition to kinesics recognition---reframes the problem as one of understanding intent. While a naïve approach might manually map each activity to a kinesic function, this becomes intractable at scale and undermines generalizability. To address this, we propose an embedded kinesics recognition framework that uses skeleton-based data to infer kinesic functions directly, bypassing the need for predefined mappings.

Our proposed framework integrates ST-GCN with CNN in a transfer learning pipeline. ST-GCN is used not to classify activities directly, but to extract latent spatiotemporal patterns from skeletal motion. These compressed representations are passed to a CNN, which learns to classify kinesic functions from this encoded structure. This approach leverages the insight that activities sharing a kinesic function tend to cluster in feature space, allowing for generalization across unseen interactions that express similar communicative intent.

We evaluated the framework across 30 randomized subsets of DUET, varying both the number and type of interactions. The results demonstrate a strong positive correlation ($\rho$=0.91, p = 0.000002) between the performance of the ST-GCN and CNN components, suggesting that improvements in representation learning are correlated with downstream kinesics recognition. This correlation supports the core hypothesis: that kinesic intent is embedded in the movement patterns themselves and can be uncovered through structured modeling. The findings affirm that DUET contains sufficient regularity and signal to support this higher-level task and establish a foundation for future research that is both semantically rich and practically scalable. Together, these metrics evaluate both components of the recognition process: representation quality (via ST-GCN accuracy and latent-space structure) and functional recoverability (via CNN accuracy), showing that the latent representations retain the information necessary for inferring communicative intent. The statistical significance also suggests that there may exist a causal relationship between ST-GCN and CNN accuracies, which requires further research investigation.

Importantly, this foundation is not an end in itself. By enabling scalable, privacy-preserving measurement of communicative function, DUET and the proposed framework equip researchers with the core capability needed to explore how specific built environment features---lighting, spatial openness, proximity of amenities---modulate interaction patterns and social signaling. In this way, our work provides the empirical and computational ``first step'' toward the human-centered design evaluation envisioned in our opening motivation.

\vspace{-0.2cm} \section{Conclusions and Future Work} \label{sec:future_work} \vspace{-0.2cm}

In this work, we introduced DUET, a contextualizable dyadic HAR dataset, together with an embedded kinesics recognition framework that operationalizes a foundational psychological taxonomy for computational sensing. These contributions jointly advance an emerging research direction: developing a measurable, theory-grounded representation of human interaction that can serve as the mechanistic layer linking social environments to social outcomes. DUET represents the first HAR dataset explicitly structured around Ekman and Friesen’s kinesics taxonomy, enabling the study of communicative function---not just physical action---within real environmental contexts. Benchmarking results demonstrate that prevailing open-source HAR algorithms struggle to capture DUET’s interactional complexity, underscoring the need for new models capable of representing dyadic, socially coordinated movement. To support scalable data collection, we additionally introduced a single-camera acquisition technique that captures multi-view variation without multi-sensor setups, enabling low-cost, flexible deployment in real-world environments.

We further proposed a kinesics recognition framework that infers communicative functions directly from skeletal movement data. Unlike survey-based or self-report methods, this framework provides a scalable and temporally responsive means of interpreting interactional intent while preserving privacy. As such, it offers a path toward sensing the interactional mechanisms that theories of social capital identify as consequential. This creates new opportunities for fields---such as urban planning, architecture, human-computer interaction, and behavioral science---that currently rely on episodic observation or self-reported experience to understand how environments shape social behavior.

Future research should advance this work along three primary directions. First, strengthening the generalizability of the kinesics recognition framework beyond DUET requires continued progress in HAR modeling. In this study, we demonstrated that kinesic communicative functions form an underlying structure embedded in HAR data, but the accuracy of function recognition remains constrained by the representational limits of current models such as ST-GCN. Because existing architectures struggle to encode dyadic and socially interactive behaviors in compact latent spaces, future work should integrate spatiotemporal models capable of more accurately capturing the relational dynamics between interacting individuals. Once such models are available, our function-based framework can be more accurately tested on external datasets to evaluate whether the same kinesic structure generalizes across even more activities and contexts. This step will further substantiate our central claim---that human movement data, when interpreted through the lens of kinesic function rather than discrete activity labels, reveal a transferable structure linking physical motion to communicative purpose. Importantly, this goal is not to construct an exhaustive dictionary of actions and meanings, but to validate the universality of function-based representations that bridge human communication theory and computational modeling. 

Second, future work should investigate how distributions of kinesic communicative functions relate to broader social outcomes—particularly social capital indicators—in real environments. Because kinesic functions correspond to interactional mechanisms such as turn-taking, emotional expression, and comfort, they offer an interpretable basis for quantifying the behavioral processes through which social infrastructure supports bonding and bridging capital. Establishing empirical links between function---level patterns and community-level outcomes would provide the evidence needed to operationalize the interaction layer as a measurable state variable in built-environment research.

Third, refining the recognition framework for deployment in situ will enable scalable, privacy-preserving sensing of interactional dynamics in real-world social infrastructure. Such advances would support continuous assessment of how physical spaces shape micro-interactions, enabling designers and planners to monitor whether a space is fostering the types of encounters theory identifies as consequential. Ultimately, embedding these behavioral indicators into design and evaluation frameworks would allow built environments to be studied—and eventually designed—through closed-loop, evidence-based approaches that center human connection and collective well-being.

\vspace{-0.2cm} \section*{Data Availability Statement} \vspace{-0.1cm} 
Some or all data, models, or code generated or used during the study are available in a repository online in accordance with funder data retention policies.

\vspace{-0.2cm} \section*{Acknowledgements} \vspace{-0.2cm} 
This work is supported by the National Science Foundation under Grant Number 2425121. \vspace{-0.2cm}

\bibliographystyle{unsrt}
\bibliography{ref}

\end{document}